\documentclass{article} 
\usepackage{iclr2027_conference,times}
\usepackage{url}
\newcommand{\net}{CoreDNA}
\usepackage{wrapfig}
\usepackage{amssymb} 
\usepackage{booktabs}
\usepackage{algorithm}
\usepackage{algpseudocode}
\usepackage{caption}
\usepackage{subcaption}
\usepackage{placeins}
\usepackage{multirow}
\usepackage{amsmath} 
\usepackage{amssymb} 
\usepackage{amsfonts}
\usepackage{bm}
\usepackage{svg}
\usepackage{siunitx}
\usepackage{cancel}
\usepackage{microtype}
\usepackage{xcolor}
\usepackage{cite}
\usepackage[export]{adjustbox}
\usepackage[hang,flushmargin]{footmisc}
\usepackage{cuted}
\usepackage[breaklinks,colorlinks]{hyperref}
\usepackage[capitalize]{cleveref}
\crefname{section}{Sec.}{Secs.}
\Crefname{section}{Section}{Sections}
\Crefname{table}{Table}{Tables}
\crefname{table}{Tab.}{Tabs.}

\definecolor{mygreen}{HTML}{00A64F}
\definecolor{myred}{HTML}{ED1B23}

\definecolor{bencolor}{HTML}{0072B2}

\definecolor{nickcolor}{HTML}{B20072}

\newcommand{\secref}[1]{Sec.~\ref{#1}}
\renewcommand{\eqref}[1]{Eq.~(\ref{#1})}
\newcommand{\figref}[1]{Fig.~\ref{#1}}
\newcommand{\tabref}[1]{Tab.~\ref{#1}}
\newcommand{\algoref}[1]{Algo.~\ref{#1}}
\newcommand{\appref}[1]{Appx.~\ref{#1}}
\iclrfinalcopy

\title{Diversity Without Labels: Task-Agnostic\\ Coreset Selection Across\\ Perception and Action}

\author{%
\begin{tabular}[t]{c}
\textbf{Julia Hindel}$^{1}$ \enspace \textbf{Benjamin Ramtoula}$^{2}$ \enspace \textbf{Kyuhwan Yeon}$^{2}$ \enspace \textbf{Nick Heppert}$^{1}$ \enspace \textbf{Paul Newman}$^{2}$ \\[8pt]
\textbf{Daniele De Martini}$^{2}$\thanks{Equal supervision. Project page: \url{coredna.cs.uni-freiburg.de}} \enspace \textbf{Abhinav Valada}$^{1}$\footnotemark[1] \\[8pt]
{\normalfont $^{1}$University of Freiburg \enspace $^{2}$University of Oxford}
\end{tabular}}

\begin{document}

\maketitle

\begin{abstract}
While data collection for model training is relatively easy, curating high-quality datasets from large collections is inherently challenging. Coreset estimation methods address this problem by identifying the best data subsets to train models on. However, existing approaches often require proxy model training, extensive annotations, or task-specific tailoring. Unsupervised alternatives typically exploit geometric properties in foundation model feature spaces, but are limited to pairwise image-to-image comparisons or comparisons against a fixed subset.
We introduce \net{}, an unsupervised, task-agnostic coreset estimation approach that pools neuron activations from vision foundation models into per-image and per-dataset distributions. By comparing each image's distribution to the dataset-level distribution, \net{} can capture both low-level visual structure and high-level semantic information across layers. We find that high-quality samples concentrate in the tails of the last layer's activation statistics, which we validate by relating them to training-based coreset methods. Building on these findings, we develop a computationally efficient selection mechanism to construct task-agnostic, high-quality coresets for selective annotation and model training. Across image classification, semantic segmentation, depth estimation, and imitation learning for end-to-end driving and mobile manipulation, \net{} matches or outperforms the state of the art, demonstrating unprecedented generalization across perception and action tasks. These results highlight the effectiveness of compact activation-based representations for data selection.\looseness=-1 
\end{abstract}


\section{Introduction}\label{sec:introduction}
Vision models have become a core component of modern perception and decision-making systems, and their performance is critically shaped by the data they are trained on.
What matters is not only how much data they see but which data, since often only a subset of a large dataset is the most informative for training.
Restricting training to such subsets can substantially improve efficiency while preserving performance, but only if the right subset can be reliably identified.
In robotics, for instance, subsets are still commonly selected based on fixed time intervals or geographic location, which neglects data diversity and might steer models toward oversampling similar scenarios~\citep{balvit25}.
As a result, training datasets are dominated by repetitive, easy-to-collect scenarios, while rare but safety-critical or informative edge cases remain underrepresented.

Coreset estimation tackles this problem by identifying a compact subset of a dataset that preserves the properties necessary for effective model training.
In particular, \emph{one-shot} coreset estimation performs this selection in a single pass, without repeatedly retraining a model to refine the coreset selection.
However, existing one-shot coreset methods face two limitations.
Some are limited to specific perception tasks, such as classification \citep{paul21el2n} or imitation learning \citep{dass2026datamil}, restricting their applicability beyond the settings for which they were designed.
Others require training on the fully-labeled dataset to determine sample "hardness," i.e., how difficult a sample is to learn, measured by its loss or misclassifications during training. This presupposes full human annotation~\citep{acharya24boss, zheng23css, killamsetty21glister}.
Nevertheless, full annotation is rarely available in practice, as human labeling is expensive and difficult to scale to large, continuously growing datasets. Unsupervised coreset estimation instead enables an a-priori data selection without relying on labels. Since diversity matters more than hardness for downstream performance~\citep{yao26div}, coreset estimation methods exploit diversity in foundation model feature spaces to yield task-agnostic, universal datasets~\citep{xie23freesel, griffin26zcore}. However, these methods score images either relative to one another, or relative to a randomly sampled reference embedding distribution, rather than relative to the dataset's own distribution. Such pairwise image-to-image comparisons are computationally expensive and do not scale to larger coresets.
In contrast, Visual DNA~\citep{ramtoula23vdna} showed that distributions of neuron activations across all layers of a frozen foundation model can be used as general-purpose representations of datasets to measure similarity, capturing both low-level and semantic attributes.
Both levels matter for task-agnostic coreset selection, as different downstream tasks rely on different cues, from appearance changes such as lighting or weather to semantic scene content.
As curating a coreset amounts to shaping the data distribution a model will be trained on, such general-purpose representations are a natural fit, and we explore how to extend and apply them to coreset selection.

To this end, we propose \net{} (Coreset DNA), an unsupervised, task-agnostic coreset estimation method that evaluates each image relative to its source dataset (\figref{fig:method}). We pool neuron activations across layers into compact per-image and dataset-level distributions and investigate the information they encode. Our analysis reveals that samples assigned high utility by training-based coreset methods tend to concentrate in the tails of final-layer activation statistics (\secref{sec:pattern}). Motivated by this, we develop computationally efficient, density and coverage-based selection criteria for selective annotation and model training. We evaluate \net{} across image classification, semantic segmentation, depth estimation, and imitation learning for end-to-end driving and mobile manipulation (\secref{sec:experiment}). 
Across these tasks, \net{} matches or outperforms state-of-the-art training-based and training-free methods, showing the value of dataset-relative activation statistics for data selection.

\section{Related Work and Background}\label{sec:related-work}
\textbf{Training-based Coreset Methods} curate coresets using signals such as losses or gradients, obtained by partially or fully training a model on the labeled dataset.
A common strategy represents each sample by the gradient of the downstream task loss on that sample with respect to the model parameters, which captures how the sample would update the model during training~\citep{baharan19craig, yang23crest, chang25grad}.
Additional works cast coreset selection as a bilevel optimization problem, with an inner loop training the model on the selected subset and an outer loop optimizing the subset itself~\citep{zhou22bilevel, killamsetty21glister, xia24bilevel}.
A separate line of work focuses on loss-based selection.
While forgetting events, i.e., switches from correct to incorrect classification between epochs, have been found to correlate with hard examples~\citep{toneva18forgetting}, AUM (Area Under the Margin) instead measures the probability gap between the target class and the largest other class across training epochs, with low values indicating hard samples~\citep{pleiss20aum}.
On the other hand, EL2N focuses on samples with a large $L2$ distance between predicted class probabilities and labels~\citep{paul21el2n}.
Follow-up work extends these methods with stratified sampling to improve data coverage at low retention rates~\citep{zheng23css}, or incorporates embedding similarity as a secondary criterion~\citep{acharya24boss, yan25coido, xie25ses, maharana24d2pruning, wu26ugies, zhang26seed} to obtain diverse coresets. 
Moreover, many methods compute selection scores with a proxy model, a cheaper model trained in place of the target one.
While denoising autoencoders select hard-to-reconstruct samples~\citep{zhang26fgbg}, smaller models from the same model family serve as cheap difficulty estimators~\citep{zhang25staff, gu2025data}. Additionally, LoRA adapters attached to large pretrained models approximate gradients efficiently~\citep{zhang24tagcos, xia24less, he26loprune, liu24tsds}.
Further methods relax the full-dataset labeling constraint. ELFS applies deep clustering for pseudo-labeling, then uses forgetting- or AUM-based scoring for coreset estimation~\citep{zheng25elfs}. OFDS extends this with object-level representations, selecting k-means centroids as the coreset~\citep{popp25ofds}.
While most training-based coreset estimation methods target image or language understanding, other related works address 
video streaming~\citep{mahdizadeh26cords}, or imitation learning~\citep{dass2026datamil}.

\textbf{Training-free Coreset Methods} instead avoid training a model on the target data to yield the coreset, and commonly rely on geometric criteria in foundation-model feature spaces to identify the most diverse samples for optimal performance.
SSL performs clustering in feature space and selects the sample closest to each cluster mean~\citep{sorscher22ssl}. Moderate~\citep{xia23moderate} instead selects data points whose distance to the class center is closest to the median distance, later improved via classwise sampling~\citep{jafari24cwerm}.
Other methods select a subset whose geometric median matches that of the full dataset~\citep{acharya25geometric}, use optimal transport to find representatives of the full dataset~\citep{zhu26oneknob}, score samples by their density in feature space after filtering those with mixed-label neighborhoods~\citep{roy26topoprune}, or maximize coverage so that every sample is close to a selected one~\citep{rustamov26mediknn}.
ZCore uses a vision foundation model to extract features, then iteratively slices the full embedding space into numerous low-dimensional subspaces, sampled via a triangular distribution to estimate each example's value based on coverage and redundancy~\citep{griffin26zcore}. FreeSel extracts representations via attention patterns from pretrained models guided by clustering and then employs distance-based greedy selection to iteratively add images dissimilar to those already selected~\citep{xie23freesel}.
TFDP ranks images for instance segmentation by a complexity score computed from annotated shapes and classes, addressing instance-area variation and class imbalance~\citep{dai25trainingtfdp}, while ViSA selects  images via foundation-model-based segmentation complexity and alignment scores~\citep{liu25visa}.
ActiveFT optimizes a parametric selection model via gradient descent on pretrained features to match the density of the unlabeled feature pool while maximizing diversity~\citep{xie223activeft}.
Further works rely on cosine similarity within a vision-language model~\citep{qi26prophet, bi26prism, maini24tmars, kamal24dbp, abbas23semdedup} or on foundation-model agreement~\citep{meding22ddd}. 
However, all outlined methods score images through pairwise image-to-image comparisons rather than capturing information about the full dataset's distribution. In this work, we examine multi-layer activation patterns relative to the full dataset, as a better proxy for the diversity of a training sample.

\section{Technical Approach} \label{sec:technical-approach}
\begin{figure*}
    \centering
    \includegraphics[width=\textwidth]{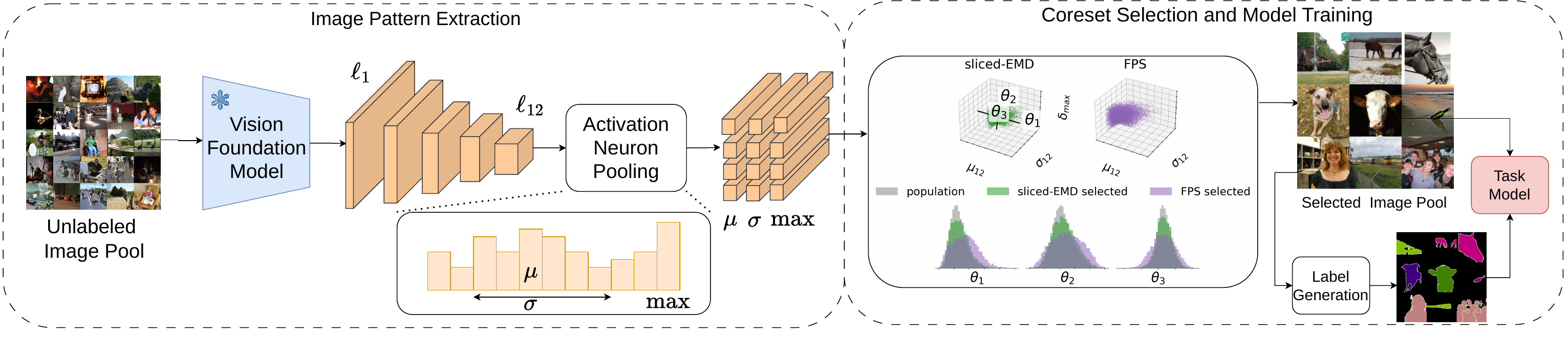} 
    \caption{\net{} extracts neuron activation histograms from a vision foundation model, pooling them into per-layer mean $\mu$, standard deviation $\sigma$, and maximum $m$. We use the last layer's $\mu$, $\sigma$, and cross-layer dispersion $\delta$ to select a coreset via sliced Wasserstein minimization (``sliced-EMD'') or farthest-point sampling (``FPS'').}
    \label{fig:method}
    \vspace{-0.3cm}
\end{figure*}

We first formalize the problem of coreset estimation and our desiderata for a suitable solution (\secref{sec:problemdefinition}). We then introduce per-image descriptors derived from pooled foundation-model activations and analyze their properties (\secref{sec:pattern}). We relate them to training-based coreset methods to uncover patterns behind good coresets (\secref{sec:selection}).
Finally, we present \net{} (\figref{fig:method}), our selection algorithm that leverages our descriptors to construct high-quality, task-agnostic coresets (\secref{sec:corednaselection}).

\subsection{Problem Definition}\label{sec:problemdefinition}
Coreset estimation selects the most informative subset of a dataset in a one-shot manner. This differs from active learning~\citep{ash20deep}, which iteratively queries an oracle for labels as training progresses.
Formally, consider a dataset $\mathcal{P}$, $|\mathcal{P}|=P$, and a fixed, pre-defined retention ratio $\alpha \in (0,1]$\footnote{Determining the optimal retention ratio is an active research direction in dataset pruning, not addressed here.}. The goal is to find a coreset $\mathcal{C} \subset \mathcal{P}$ with $|\mathcal{C}|=C=\alpha P$ that maximizes the downstream performance of a model trained on it.
Our coreset estimation method should satisfy six requirements: it should be training-free, task-agnostic, encompassing information about the full dataset's distribution, agnostic to the number and type of visual inputs, able to incorporate data recording information, and computationally scalable to larger datasets. Additionally, we aim for a compact, low-dimensional representation that allows the underlying properties to be studied in detail.
To achieve this, we divide the problem into pattern extraction and selection. Pattern extraction maps the $P$ images to a compact feature matrix $F \in \mathbb{R}^{P \times D}$, representing each image relative to the full dataset. A computationally scalable selection algorithm then selects a coreset of $C$ samples from $F$.

\begin{figure*}
    \centering
    \includegraphics[width=\textwidth]{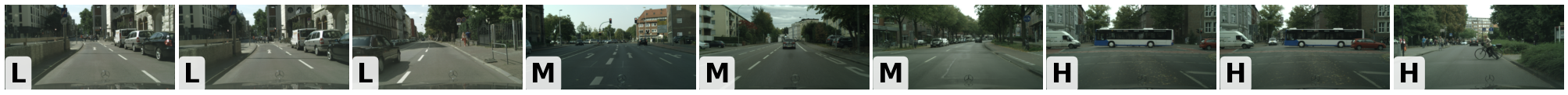} 
    \caption{Example Cityscapes-DVPS images ranked by $\mu_{12}$ into low ($L$), medium ($M$), and high ($H$) buckets. High-$\mu_{12}$ images tend to contain rare classes, while near-duplicates appear have similar scores in both $L$ and $H$.}
    \label{fig:visual}
    \vspace{-0.3cm}
\end{figure*}

\subsection{Per-Image Pattern Extraction}\label{sec:pattern}
\textbf{Image Patterns Considered:} We first explore potential ways to produce image features that might serve as a good base for our design of a coreset curation method. To do so, we first create image-wise embeddings using VDNA~\citep{ramtoula23vdna}, which was originally designed to measure dataset similarity. VDNA computes, for every neuron of a frozen ViT-B backbone, a histogram of that neuron's activations over an image $x$. We use the Mugs ViT-B/16 \citep{mugs2022SSL}, as in VDNA. The backbone has $L=12$ layers ($\ell = 1, \dots, L$), each with $N = 768$ neurons ($n = 1, \dots, N$). Neuron $(\ell, n)$ yields raw bin counts $r_{\ell n}(x)$ over $B$ bins, which we normalize into a histogram $h_{\ell n}(x) \in \Delta^{B-1}$. The dataset-level VDNA $H_{\ell n}$ is the histogram of the same neuron accumulated over the full training set. We then compute the per-neuron EMD between an image's histogram and the dataset-level histogram:
\begin{equation}
h_{\ell n}(x) = \frac{r_{\ell n}(x)}{\lVert r_{\ell n}(x) \rVert_1},
\qquad 
  H_{\ell n} = \frac{\sum_{i=1}^{P} r_{\ell n}(x_i)}
                        {\lVert \sum_{i=1}^{P} r_{\ell n}(x_i)\rVert_1},
  \qquad
  d_{\ell n}(x) = \mathrm{EMD}\big(h_{\ell n}(x),\, H_{\ell n}\big).
\end{equation}
The resulting distances $d_{\ell n}(x)$ across all neurons reflect how typical or atypical image $x$ is relative to the full dataset, at the granularity of individual neurons. To reduce dimensionality and build a compact representation, we explore multiple candidate statistics over these distances separately for each layer $\ell$. Over all the neurons $n$ of that layer $\ell$, we compute $d_{\ell n}(x)$ and record their mean $\mu_\ell(x)$, standard deviation $\sigma_\ell(x)$, and maximum $m_\ell(x)$. $\mu_\ell(x)$ captures a layer's overall atypicality, while $\sigma_\ell(x)$ and $m_\ell(x)$ reveal whether that atypicality is broadly distributed or concentrated in a few strongly reacting neurons. We omit the minimum, as it is dominated by the many near-zero, uninformative neurons and adds no additional signal. Further, we also compute the same three statistics computed over all layers jointly, $\mu_{1:L}(x)$, $\sigma_{1:L}(x)$, and $m_{1:L}(x)$. Together, these $3L + 3$ statistics form the representation $\phi(x) \in \mathbb{R}^{3L+3}$ of image $x$, yielding a $P \times (3L+3)$ representation matrix $\Phi = [\phi(x_1), \dots, \phi(x_P)]^\top$ across the full dataset, containing statistics describing diverse types of properties and visual features, some of which we hope can help identify good coresets.

\begin{wrapfigure}{R}{0.45\textwidth}
    \centering
    \vspace{-15pt}
    \includesvg[width=\linewidth]{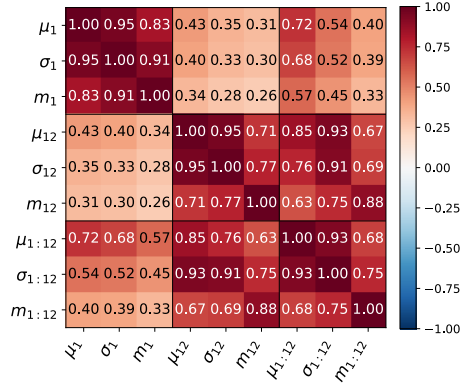}
    \caption{Correlation analysis of image patterns averaged across all datasets. Same-layer statistics correlate strongly; cross-layer correlation decays.}
    \label{fig:vdna_analysis}
    \vspace{-3pt}
\end{wrapfigure}

Beyond these per-layer statistics, we propose to explore two further candidate statistics that measure how consistently the entries of $\phi$ rank an image, to identify images where the foundation model's representation shifts across layers, hypothesizing these may add an informative axis for selection.
For each of the $3L+3$ entries, indexed by $k$, we sort all $P$ training images in ascending order of that entry and let $\rho_k(x) \in \{0, \dots, P-1\}$ denote the position of image $x$ in this ordering. We measure how much the individual representations in $\phi(x)$ disagree about the image rank. Let $\tilde{\rho}_k(x)$ denote the absolute deviation of entry $k$'s rank from the per-image median rank,
\begin{equation}
\tilde{\rho}_k(x) = \bigl|\, \rho_k(x) - \operatorname{med}_{k'} \rho_{k'}(x) \bigr|.
\end{equation}
We then define $\delta_{\max}(x) = \max_{k} \tilde{\rho}_k(x)$ and $\delta_{\mathrm{avg}}(x) = \frac{1}{3L+3}\sum_{k} \tilde{\rho}_k(x)$ as the maximum and average deviation across all $3L+3$ entries of $\phi(x)$, respectively. Small $\delta_{\max}(x)$ or $\delta_{\mathrm{avg}}(x)$ means all entries rank the image alike, whereas large values mean one or more entries rank it differently from the rest.
We add both to $\phi(x)$ as further candidate statistics, which thus contains $3L+5$ entries.

\textbf{Analysis of Image Patterns:}
We analyze the layerwise representations with the Pearson correlation coefficient $r$ of $\mu_\ell$, $\sigma_\ell$, and $m_\ell$ ($\ell = 1, 12, 1{:}12$) in \figref{fig:vdna_analysis}, with qualitative examples of $\mu_{12}$ shown in \figref{fig:visual}.
$\mu_\ell$ and $\sigma_\ell$ are strongly correlated ($r=0.93$--$0.95$) across same layers, whereas $\mu_\ell$ and $m_\ell$ show more variable correlation across datasets ($r=0.68$--$0.83$). Further, the correlation between the means of different layers decays with the distance between them and the two most distant blocks, $\ell_1$ and $\ell_{12}$, share only $r=0.43$, the weakest pair in every dataset. Even the all-layers aggregate $\mu_{1:12}$ is only moderately tied to the mean of the first and last block ($r=0.72$ for $\ell_1$, $0.85$ for $\ell_{12}$). Layers are therefore only moderately correlated on raw values, not independent, and markedly less correlated across layers than $\mu_\ell$ is with its own $\sigma_\ell$ within a layer. However, different layers' $\mu_\ell$ largely agree on which images are extreme. We compare if the top or bottom 10\% of $\mu_1$ and $\mu_{12}$ overlap and compute a Jaccard index of $0.12$--$0.27$, which is $2.3$--$5.1\times$ the chance level ($0.053$), whereas a same-size band of middle-ranked images overlaps at chance ($0.054$--$0.073$). The same images thus recur in the tails of the lowest and the highest layer, even though the layers' raw values are only moderately correlated.

Further, we compute the Pearson correlation of $\delta_{\max}(x)$ and $\delta_{\mathrm{avg}}(x)$ with $(\mu_{12}, \sigma_{12})$. We find that they are largely orthogonal: mean absolute correlation is $0.04$--$0.08$ on CIFAR10, PascalVOC, Cityscapes, Cityscapes-DVPS, and NAVSIM. There are two exceptions, SemanticKITTI-DVPS (correlation of $0.36-0.38$) and LIBERO-10 ($0.20-0.25$). In both exceptions, the correlation is negative: images with high $\mu_{12}$ (strongly atypical) tend to be ranked as atypical consistently across all layers, resulting in low $\delta$, which shows that they are semantic outliers.

\subsection{Analysis on Coreset Methods} \label{sec:selection}

\begin{wraptable}{R}{0.48\textwidth}
\vspace{-13pt}
\centering
\footnotesize
\setlength{\tabcolsep}{4pt}
\caption{Axes of the selection space (base: $\mu_{12}$, the mean of layer $\ell_{12}$). Correlation of run performance with the coverage and fidelity metrics of the selected set, averaged over the metrics of each family and over seven dataset-task combinations (5{,}265 historical runs). Best in bold.}
\label{tab:axis-summary}
\begin{tabular}{lcc}
\toprule
Selection space & Coverage $\uparrow$ & Fidelity $\uparrow$ \\
\midrule
$\mu_{12}$ & 0.273 & 0.300 \\
$\mu_{12}$ + $\delta_{\max}$ & 0.276 & 0.340 \\
$\mu_{12}$ + $\sigma_{12}$ & 0.280 & 0.341 \\
$\mu_{12}$ + $\delta_{\max}$ + $\sigma_{12}$ & \textbf{0.308} & \textbf{0.357} \\
\bottomrule
\end{tabular}
\vspace{-7pt}
\end{wraptable}

\textbf{Choice of the Selection Axes}: We introduce our dataset representation $\Phi$, which contains embeddings reflecting diverse levels of visual features and relative rarity within the full dataset.
We perform coreset selection in a low-dimensional space formed by a few of these $3L + 5$ axes to facilitate sampling, so we first aim to identify which axes form a space in which coreset quality best predicts downstream performance.
We consider candidate spaces formed by two or three axes. Since our earlier analysis showed that the top and bottom tails overlap between $\ell_1$ and $\ell_{12}$ despite their raw values being only moderately correlated, we restrict the search to combinations of statistics within a single layer, keeping the axis search space minimal while still capturing cross-layer tail agreement.
To do so, we look for patterns in 5,265 previous training runs from random, training-based, and unsupervised baselines. These span seven dataset-task combinations across five datasets and three task types (classification, segmentation, depth; 246--1,111 runs per combination).
Since downstream performance is a property of an entire coreset rather than of individual samples, relating it to our sample-wise features requires summarizing each coreset, within a given space, into a single coreset-level score.
We propose that a coreset might benefit from both covering the full dataset's support and matching its density, and we therefore consider two kinds of scores measuring each of these goals, which we call coverage and fidelity, respectively.
These two goals can conflict as matching density under-samples the tails, while maximizing coverage over-represents outliers.
We measure a coverage family (e.g., the mean distance from each sample to its nearest selected sample) and a fidelity family (e.g., the Wasserstein distance between the coreset and the full dataset), listed in \appref{sec:correlation} and oriented so that higher values indicate a better coreset.
For each candidate space, we then compute the Pearson correlation across runs between each score and downstream performance, only comparing runs with the same coreset ratio so that differences in coreset size do not drive the correlation. Then we average correlations of each family.
A high correlation indicates that coresets with high coverage (or fidelity) in that space also lead to high downstream performance, which makes it a suitable space in which to select coresets.

\textbf{Source Layer Selection for Features}: We find that the last layer's mean $\mu_{12}$, standard deviation $\sigma_{12}$, and dispersion $\delta_{\max}$ correlate best with performance on fidelity, while all candidate axes perform similarly on coverage (\tabref{tab:axis-summary} and per-dataset results in \tabref{tab:axis-second} in the appendix). A correlation of $0.357$ may seem modest at first glance, but it holds consistently across $5{,}265$ runs, spanning five datasets across three task types, each evaluated at various retention rates and with multiple selection methods. Despite training noise entirely unrelated to sample selection, this signal persists, making it a substantial finding. The two are also complementary since combining them into a three-axis space gives the highest correlations for both coverage and fidelity (0.308 and 0.357), improving over either two-axis space on most datasets (\tabref{tab:axis-third}). We note, however, that Cityscapes shows no correlation with any fidelity metric, suggesting that coverage may be the more informative metric for this dataset. Given this analysis, we represent each image $x$ by $z(x) = \bigl(\mu_{12}(x),\, \sigma_{12}(x),\, \delta_{\max}(x)\bigr)$, yielding a feature matrix $F \in \mathbb{R}^{P \times 3}$ on which we perform coreset selection.

\textbf{Additional Axis}: For tasks with multiple visual inputs, we combine them analogously, $z(x) = \bigl(z(x_a), z(x_b)\bigr)$, with $x_a$ and $x_b$ denoting the different visual inputs. Beyond images, many datasets also carry auxiliary recording information, such as timing or proprioceptive sensor readings, and we evaluate the effect of adding such non-visual information to $z(x)$. For video sequences, we add a frame index to encourage temporal diversity, since consecutive frames at high frame rates are highly redundant. For autonomous agents, we suggest incorporating pose, velocity, and acceleration. We compress this numerical information into one dimension with PCA before adding it to $z(x)$.

\begin{figure}
    \centering
    \includesvg[width=0.9\textwidth]{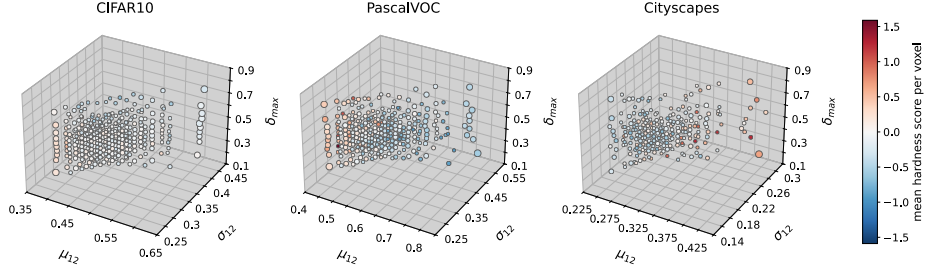}
    \caption{3D scatter of $z(x)$, colored by EL2N-measured hardness, for CIFAR10, PascalVOC, and Cityscapes.}
    \label{fig:hardness_evaluation}
   \vspace{-0.3cm}
\end{figure}

\textbf{Comparison to Training-Based Metrics}: We compare our selection features $z(x)$ to difficulty scores from a full training run such as AUM, EL2N, and Forgetting to understand where the hard samples sit along the $\mu_{12}$, $\sigma_{12}$, and $\delta_{\max}$ axes.
We report two statistics.
The first is the Pearson correlation $r$ between $\mu_{12}(x)$ and difficulty over all samples.
The second is the tail enrichment, which measures how much harder the samples in a tail of $\mu_{12}$ are than average.
We take the 10\% of samples with the highest (or lowest) $\mu_{12}(x)$ and express their mean difficulty as a $z$-score relative to the whole training set, with difficulty oriented so that positive means harder.
For example, +0.63 SD means that the average sample in that tail is 0.63 standard deviations harder than the average training sample. 
The relation differs across datasets (\figref{fig:hardness_evaluation}).
On Cityscapes, hard samples concentrate in the high-$\mu_{12}$ tail (+0.63 SD), and Cityscapes-DVPS (Depth) shows the same trend more weakly.
On PascalVOC, the relation reverses, with hard samples in the low tail (up to +0.80 SD).
On CIFAR10 and SemanticKITTI-DVPS, the relation is weak or absent.
We omit Cityscapes-DVPS (Segm.), where the difficulty scores disagree with each other.
This dataset dependence motivates offering both a coverage-based and a fidelity-based selection procedure.

\subsection{CoreDNA Coreset Selection Procedure}\label{sec:corednaselection}

Given our prior analysis of where the hardest samples lie and whether fidelity or coverage can predict it, we experiment with two  formulations: a fidelity-based one that matches the data distribution via the sliced Wasserstein distance, and a coverage-based one that spreads the coreset over the data via farthest-point sampling (FPS).
Each axis of $z(x)$ is first standardized to unit variance over the dataset so that no axis dominates the distance. We apply a redundancy heuristic such that an axis is removed when its correlation with another exceeds $0.95$, which holds for $\sigma_{12}$ and $\mu_{12}$ on Cityscapes (all variants), NAVSIM, SemanticKITTI-DVPS, and often on LIBERO-10, ensuring that the remaining axes are decorrelated and each contributes independent information to selection. 
For fidelity, we select coresets by minimizing a sliced Wasserstein distance~\citep{Kolouri19GeneralizedSW} between the coreset and the full dataset across all $D$ dimensions. In 1D, the Wasserstein distance between equal-weight empirical measures equals the mean absolute quantile difference, minimized by picking the center of each of the $P/C$ quantile bins. To extend this to $\mathbb{R}^D$, we use the sliced Wasserstein distance, drawing $E$ unit directions $\theta_1, \dots, \theta_E$, with $\theta_1$ fixed to the primary axis and the rest random. For each direction, we score each point by its rank distance to the nearest quantile center, and sum these scores across directions.
Since near-duplicates can occur, we scan points in increasing score order and skip any point within distance $g$ of an already selected one, where $g = 0.01 \cdot (0.1/\alpha)^{1/D}$ is a unit-cube spacing heuristic. If fewer than $C$ points are selected, we relax the constraint and add skipped points in score order until $C$ is reached. This selection strategy has a compute cost of $O(EP\log P)$; we show pseudocode in \appref{sec:coresettings}.

Since fidelity does not correlate with performance for some datasets (\secref{sec:selection}), we introduce farthest-point sampling (FPS) as an independent, coverage-based selection method. Starting from an arbitrary point, we repeatedly add the point farthest from the current coreset, at $O(PC)$ total cost via a running minimum-distance array. FPS is a 2-approximation for the $k$-center objective and the standard heuristic for spread-out coverage. Unlike sliced Wasserstein selection, it maximizes diversity and is therefore biased toward sparse and extreme regions.

\section{Experimental Evaluation}\label{sec:experiment}
To assess the generality of \net{}, we evaluate it across a broad range of tasks (classification, segmentation, depth estimation, driving, manipulation), multiple data domains (real, synthetic), both curated and non-curated dataset types, and varying coreset sizes. We classify our benchmarks into three main categories: curated perception datasets, non-curated perception datasets, and imitation learning. Curated datasets are off-the-shelf and already a-priori filtered without access to the raw data, consisting of single images for which all labels are available (e.g., CIFAR10). Non-curated datasets contain raw sensor observations from a sequence, such as a video with framewise labels (e.g., Cityscapes-DVPS). Imitation learning tasks are a variant of non-curated datasets with an underlying Markov decision process, which we analyze independently (e.g., NAVSIM).
We evaluate \net{} against nine baselines (\textit{Random}, \textit{Forgetting}, \textit{EL2N}, \textit{AUM}, \textit{CCS}, \textit{BOSS}, \textit{ELFS}, \textit{FreeSel}, and \textit{ZCore}) across 41 dataset-ratio settings spanning seven benchmarks (CIFAR10~\citep{krizhevsky2009learning}, PascalVOC~\citep{everingham2010pascal}, Cityscapes~\citep{cordts2016cityscapes}, Cityscapes-DVPS~\citep{qiao21dvps}, SemanticKITTI-DVPS~\citep{qiao21dvps}, NAVSIM~\citep{daunerNAVSIMDataDrivenNonReactive2024a}, and LIBERO-10~\citep{liu2023libero}). All experiments are run over three seeds, and we report averages.

\subsection{Performance On Perception Datasets}

We compare performance on CIFAR10, Cityscapes, and PascalVOC across various retention ratios $\alpha$ in \tabref{tab:main_results}. BOSS cannot be trained at $\alpha=0.01$ (24 images) since it does not function with such a small number of images. At higher retention ratios (0.5--0.7), results across all methods and datasets converge into a tight band, consistent with the intuition that keeping more data makes the choice of selection matter less. 
On CIFAR10 at low retention (0.01), our sliced-Wasserstein-based sampling outperforms all label-free baselines which we attribute to our representation's semantic sensitivity combined with density-based sampling. Purely hardness-based training methods are unsuitable at such low ratios, as their focus on difficult samples is not favorable for this task at this retention level. At higher ratios, training-dynamics-based methods perform better, while our method remains on par with other unsupervised methods while being cheaper and not requiring any additional training.
For Cityscapes, we note that FPS sampling dominates. We observe that the hardest samples lie in the high tail (\secref{sec:selection}), and oversampling these tails appears important for this task, as reflected by the high scores of training-based methods which concentrate on high samples (EL2N, AUM, and Forgetting). Only at high ratios does fidelity-based sampling slightly outperform FPS. Our method outperforms both unsupervised and training-based methods at $\alpha=0.7$ by at least $0.64$\,pp, showing that diversity-based sampling can even outperform hardness-based sampling at high retention rates.
On PascalVOC, \net{} outperforms both training-based and unsupervised methods by a margin of 2\,pp at $\alpha=0.1$, demonstrating superior selection. 
For this task as well, fidelity-based sampling is clearly better than farthest-point sampling, which we attribute to spurious outliers hurting performance under FPS. 
More broadly, we observe that neither FPS nor sliced Wasserstein sampling dominates universally, with the better choice depending on the specific dataset and task. However, \net{} with either selection consistently performs well across all settings, unlike several baselines whose performance varies substantially. In many cases, \net{} outperforms considerably stronger and more cumbersome baselines, and remains competitive even in the settings where it does not lead.

\begin{table*}[t]
\centering
\caption{Downstream accuracy (\%) across coreset ratios for CIFAR10, Cityscapes, and PascalVOC. Full: training on all data. WS: Sliced Wasserstein sampling. FPS: farthest-point-sampling.}
\label{tab:main_results}
\setlength{\tabcolsep}{3.5pt}
\resizebox{\textwidth}{!}{%
\begin{tabular}{l l ccccccc c ccccc c ccccc}
\toprule
& & \multicolumn{7}{c}{CIFAR10 $\uparrow$ (Full: 95.05)} & & \multicolumn{5}{c}{Cityscapes $\uparrow$ (Full: 76.22)} & & \multicolumn{5}{c}{PascalVOC $\uparrow$ (Full: 79.01)} \\
\cmidrule(lr){3-9} \cmidrule(lr){11-15} \cmidrule(lr){17-21}
& Method & 0.01 & 0.05 & 0.1 & 0.2 & 0.3 & 0.5 & 0.7 & & 0.01 & 0.1 & 0.3 & 0.5 & 0.7 & & 0.01 & 0.1 & 0.3 & 0.5 & 0.7 \\
\midrule
{\footnotesize \textit{Ref.}} & Random       & 41.41 & 71.21 & 80.90 & 88.17 & 90.69 & 93.11 & 94.43 & & 48.81 & 65.33 & 71.87 & 73.68 & 74.87 & & 40.02 & 55.56 & 74.81 & 76.65 & 78.06 \\
\midrule
\multirow{5}{*}{\footnotesize \shortstack[l]{\textit{Training}\\\textit{w/ labels}}} & AUM          & 16.10 & 22.80 & 31.93 & 60.54 & \textbf{92.53} & \textbf{95.24} & 95.28 & & \underline{50.96} & \underline{69.38} & 71.41 & \textbf{75.42} & 74.00 & & 26.51 & 29.03 & 71.73 & 76.84 & 78.14 \\
 & Forgetting   & 25.88 & 30.95 & 39.50 & 68.78 & 90.35 & \underline{95.08} & \underline{95.32} & & 50.42 & 69.33 & 70.61 & 74.75 & 74.68 & & 25.44 & 31.22 & 75.21 & \textbf{77.30} & \textbf{78.26} \\
 & EL2N         & 13.94 & 18.16 & 26.31 & 68.37 & 90.34 & 94.90 & \textbf{95.34} & & \textbf{51.51} & \textbf{69.88} & 71.33 & \underline{75.21} & 73.89 & & 16.38 & 29.12 & 72.93 & \underline{77.12} & \textbf{78.26} \\
 & CCS          & 41.68 & 74.88 & 86.49 & \textbf{88.85} & 91.03 & 91.16 & 91.40 & & 47.58 & 61.78 & 72.04 & 71.50 & 74.63 & & 44.01 & 43.15 & 73.94 & 75.47 & 78.05 \\
 & BOSS         & \textbf{45.75} & \textbf{78.02} & 84.28 & 87.63 & 91.43 & 94.32 & 95.33 & & -     & 64.01 & 66.82 & 69.93 & 70.11 & & \underline{46.19} & 56.44 & 75.24 & 76.75 & 77.77 \\
\midrule
\multirow{2}{*}{\footnotesize \shortstack[l]{\textit{Training}\\\textit{w/o labels}}} & ELFS AUM     & 35.21 & 74.82 & \textbf{86.66} & 87.74 & 89.21 & 89.25 & 91.13 & & 46.38 & 64.77 & 71.55 & 74.68 & 74.28 & & 45.05 & 53.00 & \textbf{75.56} & 76.93 & 77.90 \\
 & ELFS Forget  & 34.72 & \underline{76.09} & \underline{86.60} & 87.52 & 89.12 & 89.01 & 90.95 & & 46.27 & 64.96 & 71.92 & 74.37 & 73.92 & & 44.92 & 49.94 & \underline{75.45} & 76.94 & 78.00 \\
\midrule
\multirow{4}{*}{\footnotesize \shortstack[l]{\textit{Training-}\\\textit{\& label-free}}} & ZCore        & 40.60 & 66.25 & 83.56 & 87.95 & 91.43 & 93.39 & 94.45 & & 48.19 & 67.07 & 72.01 & 74.03 & 74.56 & & 39.98 & 57.52 & 75.31 & 76.76 & 78.01 \\
 & FreeSel      & 40.62 & 70.57 & 82.66 & \underline{88.26} & \underline{91.64} & 93.38 & 94.35 & & 47.98 & 64.27 & 71.90 & 74.48 & 74.68 & & \textbf{47.54} & 55.79 & 74.94 & 76.70 & 77.95 \\
 & \net{} (WS)  & \underline{41.95} & 68.61 & 82.40 & 87.95 & 91.24 & 93.49 & 94.40& & 49.92 & 66.98 & \underline{72.42} & 74.40 & \textbf{75.32} & & 44.19 & \textbf{59.21}& 75.36	& 77.09	& \underline{78.16} \\
 & \net{} (FPS) & 41.65	& 67.42		& 80.62 & 86.79	& 90.98	& 93.30	& 94.39	&& 50.20	& 	67.35	& 	\textbf{72.76} & 74.46	& \underline{75.08}	&& 37.21	& 	\underline{57.60}	& 	73.99 & 76.46	& 77.81 \\
\bottomrule
\end{tabular}%
}
\end{table*}

We now turn to non-curated, sequence-based data, where redundancy between consecutive frames introduces properties absent from curated benchmarks.
We compare Cityscapes-DVPS and SemanticKITTI-DVPS for monocular metric depth estimation and semantic segmentation in \tabref{tab:dvps_results}.
Depth estimation is a regression problem, for which the classification-specific methods \textit{AUM}, \textit{CCS}, and \textit{BOSS} cannot be applied.
SemanticKITTI-DVPS reaches within $\pm 1$\,pp of full-dataset performance at $\alpha=0.3$ for segmentation, and already at $\alpha=0.1$ for depth, showing that many samples do not meaningfully contribute to training. 
For all four non-curated datasets, training-free methods outperform training-based methods in all settings, showing that diversity-based selection is more effective than hardness-based selection when redundant data is present. We further observe that random selection outperforms training-based methods for depth estimation, indicating that these methods are limited to classification-style problems.
Further, temporal stride (i.e., selecting every $P/C$-th frame) is a strong baseline which is consistent with consecutive frames being highly uniform while sufficient diversity still emerges over time. We therefore evaluate adding a frame index as a fourth dimension (WS+T), which spreads selected frames further apart in time and outperforms pure WS in 9 of 10 segmentation settings and 2 of 6 depth settings. In general, temporal stride is difficult to beat in the mid-range ($\alpha=0.1$--$0.3$) across all four tasks, while unsupervised methods perform better at the extremes ($\alpha=0.01$ and $\alpha \geq 0.5$). \net{} performs consistently well across this entire range, and particularly excels at high retention ratios on Cityscapes-DVPS segmentation, outperforming baselines by at least $0.82$\,pp at $\alpha \geq 0.5$.


\begin{table*}
\centering
\caption{Downstream accuracy across coreset ratios for Cityscapes-DVPS and SemanticKITTI-DVPS. Segmentation is evaluated in mIoU [\%] and monocular metric depth estimation in RMSE. Full: training on all data. WS: Sliced Wasserstein sampling. FPS: farthest-point-sampling.}
\label{tab:dvps_results}
\setlength{\tabcolsep}{3.5pt}
\resizebox{\textwidth}{!}{%
\begin{tabular}{l l ccccc c ccc c ccccc c ccc}
\toprule
& & \multicolumn{5}{c}{\shortstack{Cityscapes-DVPS\\Segm. $\uparrow$ (Full: 66.43)}} & & \multicolumn{3}{c}{\shortstack{Cityscapes-DVPS\\Depth $\downarrow$ (Full: 3.91)}} & & \multicolumn{5}{c}{\shortstack{SemanticKITTI-DVPS\\Segm. $\uparrow$ (Full: 55.58)}} & & \multicolumn{3}{c}{\shortstack{SemanticKITTI-DVPS\\Depth $\downarrow$ (Full: 3.39)}} \\
\cmidrule(lr){3-7} \cmidrule(lr){9-11} \cmidrule(lr){13-17} \cmidrule(lr){19-21}
& Method & 0.01 & 0.1 & 0.3 & 0.5 & 0.7 & & 0.01 & 0.1 & 0.3  & & 0.01 & 0.1 & 0.3 & 0.5 & 0.7 & & 0.01 & 0.1 & 0.3 \\
\midrule
\multirow{2}{*}{\footnotesize \textit{Ref.}} & Temporal Stride & \underline{38.28}	& \textbf{56.71}	& \textbf{63.83}	& \underline{64.47}	& 65.02	& & 6.17	& \textbf{4.28}	& \textbf{3.97}	& & 38.74	& \textbf{53.14}	& \textbf{55.28}	& 55.80	& 56.32 & & \underline{3.67}	& \textbf{3.34}	& 3.35\\
 & Random          & 37.88 & 55.77 & 61.23 & 64.03 & 64.83 & & 5.70 & 4.36 & \underline{4.02} & & 36.81 & 51.17 & 55.08 & 55.98 & 55.90 & & 3.70 & 3.36 & 3.36 \\
\midrule
\multirow{5}{*}{\footnotesize \shortstack[l]{\textit{Training}\\\textit{w/ labels}}} & AUM             & 35.54 & 51.28 & 57.52 & 60.56 & 63.21 & & -    & -    & -      & & 27.76 & 42.85 & 51.03 & 53.45 & 55.06 & & -    & -    & -     \\
 & Forgetting      & 32.86 & 45.70 & 56.63 & 60.15 & 62.63 & & \underline{5.56} & 4.44 & 4.12 & & 26.44 & 40.35 & 46.26 & 52.84 & 54.46 & & 3.95 & 3.45 & \underline{3.34} \\
 & EL2N            & 32.84 & 52.01 & 58.99 & 62.00 & 64.65 & & 7.40 & 5.74 & 5.15 & & 28.07 & 43.58 & 51.12 & 53.55 & 55.04 & & 3.74 & 3.52 & 3.40 \\
 & CCS             & 37.98 & 50.43 & 58.15 & 60.53 & 62.16 & & -    & -    & -  & & 27.76 & 50.63 & 54.69 & 55.18 & 55.37 & & -    & -    & -     \\
 & BOSS            & 34.47 & 51.13 & 57.53 & 57.70 & 58.39 & & -    & -    & -     & & 38.07 & 51.04 & 52.93 & 52.30 & 54.35 & & -    & -    & -     \\
\midrule
\multirow{2}{*}{\footnotesize \shortstack[l]{\textit{Training}\\\textit{w/o labels}}} & ELFS AUM        & 36.53 & 52.77 & 59.60 & 62.28 & 64.89 & & 6.28 & 4.36 & 4.07 & & 38.23 & 46.96 & 50.82 & 52.63 & 54.27 & & 3.77 & 3.42 & 3.37  \\
 & ELFS Forget     & 36.65 & 52.64 & 59.79 & 62.76 & 64.18 & & 6.29 & 4.36 & 4.07 & & 38.31 & 47.31 & 54.05 & 55.65 & 55.74 & & 3.74 & 3.43 & 3.39  \\
\midrule
\multirow{5}{*}{\footnotesize \shortstack[l]{\textit{Training-}\\\textit{\& label-free}}} & ZCore           & 38.07 & 55.05 & 61.54 & 63.33 & 64.30 & & 5.81 & 4.43 & 4.08 & & 38.69 & \underline{52.47} & 54.89 & 54.90 & \underline{56.24} & & \textbf{3.60} & 3.39 & 3.37  \\
 & FreeSel         & 38.12 & \underline{56.30} & \underline{63.18} & 63.36 & 64.40 & & \textbf{5.02} & \underline{4.31} & \textbf{3.97} & & \textbf{39.33} & 51.74 & \underline{55.16} & \textbf{56.22} & 55.90 & & 3.74 & \underline{3.35} & \textbf{3.33}  \\
 & \net{} (WS) & 38.18 & 55.66	& 61.70	& 64.00 & 64.52 & & 6.05 &	4.35	& 4.05 &	& 37.56	& 50.02	& 54.90	& 55.50 &	55.62	& &  3.73	& 3.39 & 	3.35 \\
 & \net{} (WS+T) & \textbf{38.51} &	55.38 & 	62.04	& \textbf{64.77}	& \textbf{65.22}	& & 6.10 & 	4.39& 	4.08 & & 37.73	& 51.37	& \underline{55.16}	& \underline{55.99} &	\textbf{56.39}& 	& 3.77& 	3.36& 	3.35	\\
 & \net{} (FPS) & 37.36	& 53.79	& 61.09	& 63.26	& \underline{64.90}	& & 6.20	& 4.40	& 4.06 & & \underline{38.89}	& 49.50	& 54.05	& 55.65	& 55.74	& & 3.74	& 3.41	& 3.40	 \\
\bottomrule
\vspace{-0.8cm}
\end{tabular}}
\end{table*}

\subsection{Performance on Imitation Learning}
To demonstrate \net{}'s applicability, we apply it to end-to-end driving and visuo-motor policy learning. Since these policies process frames independently, we apply \net{} at the frame level.

\textbf{Autonomous Driving}:
For end-to-end driving, we complement the features from images with the ego \emph{state}, describing how the vehicle moves (velocity, acceleration, and heading), and its \emph{position}, describing where the vehicle is in global map coordinates.
We report results using the common Extended Predictive Driver Model Score (EPDMS) of Latent TransFuser~\citep{chittaTransFuserImitationTransformerBased2023} on the NAVSIM framework in \tabref{tab:navsim}. With farthest-point sampling, selecting 50\% of the data nearly matches training on the full dataset. Combining the image representation with both state and position performs best at $\alpha=0.1$, whereas at $\alpha=0.5$, excluding position from the selection is preferable. We attribute this to $\alpha=0.1$ benefiting from explicitly selecting for diverse locations at such a low sample count, whereas at $\alpha=0.5$, sufficient location diversity is already achieved naturally, making the position axis redundant. In general, FPS outperforms WS and the random baseline at both ratios, indicating that picking rare samples is helpful for this task, and \net{} can significantly improve autonomous driving policy training efficiency with minimal performance drop.

\begin{wraptable}{R}{0.4\textwidth}
\centering
\vspace{-13pt}
\caption{Downstream driving performance on NAVSIM. Img, State, and Pos. indicate the coreset selection axes.}
\label{tab:navsim}
\resizebox{0.4\textwidth}{!}{\setlength{\tabcolsep}{2pt}
\begin{tabular}{lcccc c}
\toprule
 & & & & \multicolumn{2}{c}{EPDMS $\uparrow$} \\
\cmidrule(lr){5-6}
Method & Img & State & Pos. & 0.1 & 0.5 \\
\midrule
Full     &  & & & \multicolumn{2}{c}{80.51}  \\
Random   &  & & & 65.44 &  78.88 \\
\midrule
\net{} (WS)                & \checkmark &      & & 66.31      &       78.17   \\
\net{} (WS)        & \checkmark & \checkmark &  &  65.29         &  78.90  \\
\net{} (WS) & \checkmark & \checkmark & \checkmark & \underline{67.03} & 78.62 \\
\net{} (FPS)                      & \checkmark &            &            & 66.98 & 78.60 \\
\net{} (FPS)              & \checkmark & \checkmark &    & 65.42        & \textbf{79.98} \\
\net{} (FPS)       & \checkmark & \checkmark & \checkmark & \textbf{67.43} & \underline{79.76} \\
\bottomrule
\end{tabular}}
\vspace{-7pt}
\end{wraptable}

\textbf{Visuo-Motor Policies}: %
\begin{wraptable}{R}{0.4\textwidth}
\centering
\vspace{-13pt}
\caption{Mean success rate across all tasks in LIBERO-10.}
\label{tab:libero}
\resizebox{0.4\textwidth}{!}{
\begin{tabular}{lccc}
\toprule
 & External & Wrist & Success \\
Method & Img & Img & Rate $\uparrow$ \\
\midrule
\multicolumn{4}{c}{\textit{LIBERO-10 Frame-level sampling}} \\
Full (5 demos) & & & 16.9 \\
\net{} (WS) & \checkmark & & \textbf{17.3} \\
\net{} (WS) & \checkmark & \checkmark & 16.8 \\
\net{} (FPS) & \checkmark & \checkmark & \underline{17.1} \\
\midrule
\multicolumn{4}{c}{\textit{LIBERO-10 Frame-level sampling (50 demos)}} \\
Full (50 demos) & & & 76.7 \\
\net{} (WS) & \checkmark & \checkmark & \textbf{77.8} \\
\midrule
\multicolumn{4}{c}{\textit{LIBERO-90 Demonstration-level sampling (mix-in)}} \\
All & & & 18.3 \\
DataMIL & \checkmark & \checkmark & \underline{25.3} \\
\net{} (WS) & \checkmark & \checkmark & 19.0 \\
\net{} (CR) & \checkmark & \checkmark &  \textbf{27.3} \\
\bottomrule
\end{tabular}}
\vspace{-7pt}
\end{wraptable}
For visuo-motor policy learning, we demonstrate the capabilities of \net{} on all 10 available tasks of the LIBERO-10 benchmark, which comprises long-horizon manipulation tasks. We perform two experiments aligned with DataMIL's~\citep{dass2026datamil} 5-demonstration setup, i.e. the base for both experiments includes a set of 5 out of the 50 available demonstrations for each LIBERO-10 task. We use a retention rate $\alpha=0.9$ for all experiments. We show major results in \tabref{tab:libero} and full results in \secref{app:libero:results}. 
In the first experiment, we select frames from the pooled dataset of all selected 5 demonstrations of the LIBERO-10 task and extend and ablate \net{}'s formulation against combinations of the external camera image and the agent's wrist camera image. 
Additionally, we show results when performing frame selection on all 50 LIBERO-10 demonstrations. We observe that \net{} improves over or is on-par with full frame sampling for both the 5 and 50 demonstration variant. We note that the retention rate itself may be suboptimal, and other values could yield even better results. Nonetheless, these results demonstrate that \net{} can improve performance over naively training using all available frames.


In our second experiment, we follow DataMIL's experimental setup and keep all 5 demonstrations fixed and rather select an auxiliary subset of 10\% demonstrations from the full LIBERO-90 demonstration corpus to mix with the 5 LIBERO-10 demonstrations during training. Thus, the focus shifts as the goal is no longer to select the most diverse frames from a training set but rather the best auxiliary mix-in demos. As a result, we alter our original formulations (WS and FPS) and select demonstrations from the LIBERO-90 dataset which 
rank closest to the five LIBERO-10 demonstrations, rather than matching LIBERO-90's own distribution. We denote this variant \net{} (CR). In this experiment, we observe a stark increase in the success rate, aligned with current observations regarding data composition for imitation learning~\citep{hejna2024remix}. This shows that pooling activations across a demonstration's frames also yields a good representation of sample quality for imitation learning, which even outperforms the training-based method DataMIL~\citep{dass2026datamil}.

\section{Conclusion}\label{sec:conclusion}
In this work, we investigated pooling neuron activations into per-image and per-dataset distributions for one-shot, task-agnostic coreset estimation. We find that the last layer's mean, standard deviation, and cross-layer rank agreement correlate best with downstream performance, which we attribute to hard samples concentrating in the tails of their distribution. Further, we introduced recording priors, such as frame index or position, as an additional axis to address redundancy in non-curated datasets.
We employed farthest-point and sliced Wasserstein sampling to construct high-quality coresets. We demonstrated that our selection is task- and data-type-agnostic on image classification, semantic segmentation, depth estimation, and imitation learning, outperforming most baselines while remaining computationally efficient.
We believe these findings can help extend coreset estimation beyond curated datasets and toward real-world action domains.
\subsection*{AI use statement}


In this work, we used generative AI tools for writing assistance and implementation of methods. We have not used generative AI tools for related work discovery or for generating research ideas. We have reviewed all AI-assisted work. We take responsibility for the final content of this work, including text, claims, or artifacts produced with the aid of generative AI.






\subsection*{Reproducibility statement}

We include implementation details in Section A.2 and A.3 in the Appendix. We make the code publicly available upon acceptance.






\footnotesize
\bibliography{references}
\bibliographystyle{iclr2027_conference}

\newpage
\appendix
\section{Appendix}

\subsection{Overview}
\begin{wrapfigure}{r}{0.5\linewidth}
\centering
\vspace{-13pt}
\begin{minipage}{\linewidth}
\scriptsize
\begin{algorithmic}[1]
\Require Points $F=\{z_i\}_{i=1}^{P}\subset[0,1]^{D}$, budget $C$, projections $E$, minimum gap $g$
\Ensure Selected index set $S\subseteq\{1,\dots,P\}$ with $|S|=C$
\State Sample $E-1$ random unit directions $\theta_2,\dots,\theta_E$; set $\theta_1 \gets e_1$
\State $w \gets P/C$; \quad $s_i \gets 0$ for all $i$
\For{$e = 1,\dots,E$}
    \State Rank all points by $\theta_e^{\top} z_i$, giving ranks $\rho_i \in \{0,\dots,P-1\}$
    \State $s_i \gets s_i + \min_j \big|\, \rho_i - (j+\tfrac12)w \,\big|$
\EndFor
\State $\pi \gets$ indices sorted by $s_i$ ascending; $S \gets \emptyset$
\For{$i \in \pi$ \textbf{while} $|S| < C$}
    \If{$z_i$ is at least $g$ from every point already in $S$}
        \State $S \gets S \cup \{i\}$
    \EndIf
\EndFor
\State \Return $S$, filling any remaining slots by score if $g$ prevented reaching $C$
\end{algorithmic}
\captionof{algorithm}{Sliced Wasserstein coreset selection}
\label{alg:sliced-emd-short}
\end{minipage}
\vspace{-7pt}
\end{wrapfigure}
We describe additional \net{} selection settings in \secref{sec:coresettings}, and the experimental setup for our tasks in \secref{sec:trainsetting}. We then show extended experimental results for end-to-end driving in \secref{sec:results-navsim} and visuo-motor policy learning in \secref{app:libero:results}. We also expand on our performance correlation experiments from the paper in \secref{sec:correlation}, detailing selection criteria and per-dataset results. We then compare the rankings of image patterns produced by different backbones and justify our choice of histogram bin size in \secref{sec:vdna_bb}. 
\secref{sec:axis-choice} shows ablations of our axis choices and selection mechanism. Finally, we show that training-based methods struggle with redundancy, using an artificial setup of duplicated CIFAR10 in \secref{sec:cifardub}.

\subsection{Additional Hyperparameters and Pseudo-Code}\label{sec:coresettings}
We use $B=16$ bins to produce the per-image histograms $h_{\ell n}$ and per-dataset histograms $H_{\ell n}$, and ablate this choice in \secref{sec:vdna_bb}. For the sliced Wasserstein distance, we set the number of projections $E=10$ and ablate additional settings in \secref{sec:axis-choice}. We additionally show the pseudocode of our sliced Wasserstein selection algorithm in \algoref{alg:sliced-emd-short}.

\subsection{Detailed Experimental Setting}\label{sec:trainsetting}
We evaluate our method on image classification, semantic segmentation, metric depth estimation, end-to-end autonomous driving and imitation learning. We use the authors' published code for all baselines. We train all experiments over $3$ seeds to remove the effect of training-specific variations. We train all coresets for a dataset with consistent numbers of iterations. We detail the set-up for each below. 

\paragraph{Image classification.} 
Following prior work~\citep{zheng23css, xie23freesel, zheng25elfs}, we use a ResNet18~\citep{he2016deep} architecture to train a network on CIFAR10~\citep{krizhevsky2009learning} with Cross Entropy loss. We train for $40{,}000$ iterations with a learning rate of $0.1$ using an SGD optimizer with momentum $0.9$, weight decay of $5\mathrm{e}{-4}$, and cosine annealing learning rate. We apply a random $32\times32$ crop with 4-pixel reflect padding and a random horizontal flip, followed by mean and standard-deviation normalization. We report results in terms of accuracy.

\paragraph{Semantic Segmentation.}
We perform semantic segmentation on PascalVOC~\citep{everingham2010pascal}, Cityscapes~\citep{cordts2016cityscapes}, Cityscapes-DVPS~\citep{qiao21dvps}, and SemanticKITTI-DVPS~\citep{qiao21dvps}.
We train a DeepLabv3~\citep{chen17dl} with a ResNet-101 backbone and a batch size of $16$ with Cross Entropy loss. We use a learning rate of $0.03$ for PascalVOC, $0.1$ for Cityscapes and SemanticKITTI-DVPS, and $0.3$ for Cityscapes-DVPS. We set the backbone's learning rate to $0.1\times$ the head's. We train for $39{,}720$ iterations on PascalVOC, $11{,}160$ on Cityscapes, $9{,}000$ on Cityscapes-DVPS, and $23{,}920$ on SemanticKITTI-DVPS, corresponding to $60$ epochs on the full training set for PascalVOC, Cityscapes, and Cityscapes-DVPS, and $30$ epochs for SemanticKITTI-DVPS. All four segmentation datasets share a random horizontal flip ($p=0.5$) and ImageNet-statistics normalization as base augmentations. PascalVOC and Cityscapes-DVPS (Segm.) use a random resized crop (PascalVOC: $513\times513$, scale $[0.5, 2.0]$; Cityscapes-DVPS (Segm.): $512\times512$, scale $[0.5, 2.0]$) with no color augmentation. Cityscapes additionally applies color jitter (brightness, contrast, saturation $=0.5$) on top of a $768\times768$ random resized crop. SemanticKITTI-DVPS (Segm.) instead resizes to $370\times1226$ and takes a random $370\times784$ crop, combined with milder color jitter (brightness, contrast, saturation, hue $=0.1$).
We use an SGD optimizer with a polynomial learning rate schedule. We report mean Intersection over Union (mIoU). To extend \textit{Forgetting}, \textit{EL2N}, and \textit{AUM} beyond classification, we define correctness as at least 80\% correctly classified pixels for segmentation.

\paragraph{Metric Depth estimation.}
We perform metric depth estimation on SemanticKITTI-DVPS and Cityscapes-DVPS. We train a ZoeDepth-Small network~\citep{zoedepth23} with a Depth Anything V2 backbone~\citep{yang24da}, using a scale-invariant log loss with a weight of $10$. We train for $23{,}920$ iterations on SemanticKITTI-DVPS and $4{,}500$ on Cityscapes-DVPS (amounting to $30$ epochs on the full dataset), using a batch size of $16$, a learning rate of $2\mathrm{e}{-5}$ for the head, and $4\mathrm{e}{-7}$ for the encoder. Both depth datasets share a heavier augmentation pipeline than their segmentation counterparts: a random horizontal flip ($p=0.5$), a random affine transform (rotation $\pm2.5^\circ$, translation $5\%$, scale $[0.95, 1.05]$), color jitter (brightness, contrast, saturation, hue $=0.1$), and random gamma correction (range $[0.8, 1.2]$), followed by ImageNet-statistics normalization. Cityscapes-DVPS uses a random resized crop to $518\times518$ (scale $[0.5, 2.0]$), while SemanticKITTI-DVPS resizes to $350\times1204$ and takes a random $350\times518$ crop.
We use an Adam optimizer with a cosine annealing learning rate scheduler. We report Root Mean Squared Error (RMSE) as the primary metric. For depth estimation, we define correctness as an absolute relative error below $0.1$ for at least $10\%$ of the image, a threshold determined empirically.

\paragraph{Imitation Learning -- Driving.}
We perform end-to-end driving on the NAVSIM benchmark. We train on the \texttt{navtrain} split, which includes both training and validation scenes, and evaluate on the held-out \texttt{navtest} split. For the baseline model, we use Latent TransFuser~\citep{chittaTransFuserImitationTransformerBased2023}, which uses only camera inputs without LiDAR. We train for $50$ epochs with a batch size of $128$ and a learning rate of ${1e-4}$. We refer the reader to the original work for further training details~\citep{cao2025pseudosimulation, chittaTransFuserImitationTransformerBased2023}. We report the Extended Predictive Driver Model Score (EPDMS) as the primary metric~\citep{daunerNAVSIMDataDrivenNonReactive2024a, cao2025pseudosimulation}.

\paragraph{Imitation Learning -- Manipulation.} \label{app:libero:experimentalsetup}

We perform end-to-end closed-loop visuo-motor policy training on the LIBERO-10 benchmark~\citep{liu2023libero} using the OCTO model~\citep{octo_2023}. We fine-tune OCTO using off-the-shelf parameters following the DataMIL benchmark~\citep{dass2026datamil} configuration, thus we use an action chunking head of size eight, a batch size of 128 and an annealing cosine learning rate schedule with an initial learning rate of $3e-4$. We evaluate all trained policies on the LIBERO-10 benchmark for 50 evaluation roll-outs and report the average success rate over them.

For selecting the five demonstration subset of all 50 demonstrations for each LIBERO-10 task, we could not rely on DataMIL's~\citep{dass2026datamil} selection as it was unavailable to us. Instead, we first filter all failed demonstrations by replaying each action sequence from the initial simulator state and only keep the ones that are successful. Second, we manually go through the remaining demonstrations and select the most diverse and relevant ones, as a human would if they had to collect them, knowing one has a budget of 5 demonstrations. For future reproducibility we report all selected demonstration indices in \tabref{libero10:demos}.

We further include the robot's proprioception (joint states) through PCA projection in the axis selection in \secref{app:libero:results}.

\begin{wraptable}{r}{0.4\textwidth}
\centering
\caption{Selected demonstration indices for each LIBERO-10 task.}
\label{libero10:demos}
\small
\begin{tabular}{ll}
\toprule
Task & Demos \\
\midrule
Book--Caddy      & \{7, 8, 19, 27, 41\}   \\
Bowl--Cabinet    & \{11, 26, 32, 38, 43\} \\
Mug--Microwave   & \{23, 27, 37, 44, 48\} \\
Stove--Moka      & \{19, 27, 30, 40, 43\} \\
Soup--Sauce      & \{3, 11, 25, 44, 45\}  \\
Cream--Butter    & \{8, 13, 29, 36, 42\}  \\
Mug--Mug         & \{9, 12, 16, 17, 23\}  \\
Mug--Pudding     & \{4, 5, 11, 12, 31\}   \\
Soup--Cheese     & \{1, 2, 22, 23, 44\}   \\
Moka--Moka       & \{7, 9, 13, 20, 44\}   \\
\bottomrule
\end{tabular}
\end{wraptable}

\subsection{Detailed Performance on End-to-End Driving}\label{sec:results-navsim}

EPDMS~\citep{daunerNAVSIMDataDrivenNonReactive2024a, cao2025pseudosimulation} consists of No at-fault Collisions (NC), Drivable Area Compliance (DAC), Driving Direction Compliance (DDC), Traffic Light Compliance (TLC), Ego Progress (EP), Time to Collision (TTC), Lane Keeping (LK), History Comfort (HC), and Extended Comfort (EC). For each scenario, the score is computed as
\begin{equation}
  \mathrm{EPDMS} = \prod_{m \in \mathcal{M}_{\text{mult}}} \tilde{m} \;\cdot\; \frac{\sum_{m \in \mathcal{M}_{\text{wt}}} w_m  \tilde{m}}{\sum_{m \in \mathcal{M}_{\text{wt}}} w_m},
  \label{eq:epdms}
\end{equation}
where $\mathcal{M}_{\text{mult}} = \{\mathrm{NC}, \mathrm{DAC}, \mathrm{DDC}, \mathrm{TLC}\}$ are multiplier metrics and $\mathcal{M}_{\text{wt}} = \{\mathrm{EP}, \mathrm{TTC}, \mathrm{LK}, \mathrm{HC}, \mathrm{EC}\}$ are weighted metrics with $w = \{5, 5, 2, 2, 2\}$. $\tilde{m}$ denotes the score of the driving model after false-positive filtering, which sets the score to $1$ whenever the human driver also violates the same metric in that scenario. The final EPDMS is averaged over all scenarios.

Table~\ref{tab:navsim_full} reports all sub-metrics. Reducing the training data primarily affects DAC and EC, whereas it has little effect on TLC and HC. TLC and HC are nearly saturated even with $10\%$ of the data. In contrast, DAC, which requires precise perception of road boundaries, and EC, which is sensitive to noisy trajectory predictions, degrade markedly with less data. Compared to random selection, at $\alpha=0.1$, the best configuration, \net{} (FPS) with image, state, and position, improves on the safety-related metrics NC ($+1.1$), DAC ($+1.5$), and TTC ($+1.4$), as well as on EC ($+3.6$), while EP is slightly lower ($-0.4$). At $\alpha=0.5$, the best configuration, \net{} (FPS) with image and state, again gains mainly on DAC ($+1.1$), with smaller improvements on NC, DDC, and TTC, while EP is again slightly lower ($-0.7$). In both cases, \net{} thus trades a small amount of progress for safer driving behavior.

\begin{table}[t]
\centering
\caption{Per-metric end-to-end driving performance on NAVSIM.}
\label{tab:navsim_full}
\resizebox{\textwidth}{!}{\setlength{\tabcolsep}{3pt}
\begin{tabular}{lcccccccccccccc}
\toprule
Method & Img & State & Pos. & Ratio & NC $\uparrow$ & DAC $\uparrow$ & DDC $\uparrow$ & TLC $\uparrow$ & EP $\uparrow$ & TTC $\uparrow$ & LK $\uparrow$ & HC $\uparrow$ & EC $\uparrow$ & EPDMS $\uparrow$ \\
\midrule
Full &  &  &  & -- & 96.99 & 89.20 & 99.07 & 99.74 & 87.81 & 95.86 & 95.50 & 98.34 & 85.95 & 80.51 \\
\midrule[\heavyrulewidth]
Random &  &  &  & 0.1 & 93.23 & 78.27 & 94.31 & 99.65 & 86.27 & 91.22 & 88.61 & 98.27 & 72.67 & 65.44 \\
\cmidrule{1-15}
\net{} (WS) & \checkmark &  &  & 0.1 & 93.64 & 78.88 & 94.69 & 99.64 & 86.20 & 92.01 & 89.95 & 98.28 & 67.53 & 66.31 \\
\net{} (WS) & \checkmark & \checkmark &  & 0.1 & 93.03 & 77.97 & 94.01 & 99.60 & \textbf{86.81} & 91.38 & 88.58 & 98.28 & 73.32 & 65.29 \\
\net{} (WS) & \checkmark & \checkmark & \checkmark & 0.1 & 93.69 & 79.31 & 94.14 & 99.65 & 86.01 & 92.10 & 89.24 & 98.27 & \textbf{77.09} & 67.03 \\
\net{} (FPS) & \checkmark &  &  & 0.1 & 93.60 & 79.09 & \textbf{94.91} & 99.65 & 86.22 & 92.03 & 90.35 & 98.28 & 74.04 & 66.98 \\
\net{} (FPS) & \checkmark & \checkmark &  & 0.1 & 93.18 & 77.53 & 94.24 & 99.64 & 85.97 & 91.26 & \textbf{90.53} & 98.29 & 73.80 & 65.42 \\
\net{} (FPS) & \checkmark & \checkmark & \checkmark & 0.1 & \textbf{94.29} & \textbf{79.77} & 94.20 & \textbf{99.66} & 85.87 & \textbf{92.61} & 89.28 & \textbf{98.31} & 76.31 & \textbf{67.43} \\
\midrule[\heavyrulewidth]
Random &  &  &  & 0.5 & 96.73 & 88.02 & 98.17 & 99.72 & \textbf{87.71} & 95.43 & 94.41 & 98.31 & 86.83 & 78.88 \\
\cmidrule{1-15}
\net{} (WS) & \checkmark &  &  & 0.5 & 96.62 & 87.67 & 97.97 & 99.69 & 87.61 & 95.16 & 92.83 & 98.33 & 86.14 & 78.17 \\
\net{} (WS) & \checkmark & \checkmark &  & 0.5 & 96.67 & 88.02 & 98.29 & 99.73 & 87.25 & 95.54 & 93.79 & 98.33 & 86.64 & 78.90 \\
\net{} (WS) & \checkmark & \checkmark & \checkmark & 0.5 & 96.50 & 87.96 & 98.37 & 99.73 & 87.51 & 95.21 & 94.21 & 98.32 & 86.21 & 78.62 \\
\net{} (FPS) & \checkmark &  &  & 0.5 & 96.58 & 87.85 & 98.04 & 99.71 & 87.69 & 95.44 & 94.04 & 98.31 & 85.93 & 78.60 \\
\net{} (FPS) & \checkmark & \checkmark &  & 0.5 & \textbf{96.93} & \textbf{89.11} & \textbf{98.57} & 99.73 & 87.05 & 95.65 & \textbf{94.51} & \textbf{98.34} & \textbf{86.98} & \textbf{79.98} \\
\net{} (FPS) & \checkmark & \checkmark & \checkmark & 0.5 & 96.85 & 88.90 & 98.23 & \textbf{99.74} & 87.58 & \textbf{95.72} & 93.86 & 98.31 & 86.28 & 79.76 \\
\bottomrule
\end{tabular}}
\end{table}

\subsection{Detailed Performance for End-to-End Manipulation Imitation Learning} \label{app:libero:results}
We show full task results in \tabref{tab:liberofull}. In addition to the findings in the main section of the paper, we observe that the full method only wins in four out of ten tasks for 5 demonstrations and in three out of ten tasks for 50 demonstrations, following the trend of the mean success rate across all tasks. The same observation can also be made for the LIBERO-90 mix-in experiment, where only on one task the full set is winning. Further, Data-MIL~\citep{dass2026datamil} shows the best performance on three out of ten tasks. 

We attribute the better/on-par performance to the fact that continuously recording during task demonstration will produce irrelevant frames that can be pruned. We would expect an even higher success rate, when extending \net{} to also include different modalities like force as it would provide additional information for frames in which images are mostly uninformative which is during the most crucial part of manipulation: dexterous interaction with the environment. We are excited about this future research direction to extend coreset estimation methods to robot manipulation.

Unlike the autonomous driving scenario, where adding low-dimensional states (i.e., joint states) improved performance, we do not observe a similar increase here.

\begin{table*}[t]
\centering
\caption{Mean success rate across all tasks in LIBERO-10, with per-task results, split across all experiments. We report different variations of \net{}'s modality input, the external camera image, the agent's wrist camera image and its joint state.}
\label{tab:liberofull}
\resizebox{\textwidth}{!}{
\begin{tabular}{lccc cccccccccc c}
\toprule
 & Ext & Wrist & Joint & Soup-- & Cream-- & Stove-- & Bowl-- & Mug-- & Book-- & Mug-- & Soup-- & Moka-- & Mug-- & Success \\
Method & Img & Img & State & Sauce & Butter & Moka & Cabinet & Mug & Caddy & Pudd. & Cheese & Moka & Micr. & Rate $\uparrow$ \\
\midrule
\multicolumn{15}{c}{\textit{LIBERO-10 Frame-level sampling}} \\
Full (5 demos) & & & 
    & 9.3\tiny$\pm$5.0 & \textbf{25.3}\tiny$\pm$11.6 & 30.7\tiny$\pm$9.0 & 36.7\tiny$\pm$16.7 & \textbf{6.0}\tiny$\pm$7.1 & 17.3\tiny$\pm$3.4 & 1.3\tiny$\pm$1.9 & \textbf{14.0}\tiny$\pm$2.8 & \textbf{10.0}\tiny$\pm$6.5 & 18.7\tiny$\pm$6.2 & 16.9\tiny$\pm$7.0 \\
\net{} (WS) & \checkmark & & 
    & 16.0\tiny$\pm$11.8 & 21.3\tiny$\pm$5.2 & 32.0\tiny$\pm$2.8 & 40.0\tiny$\pm$5.9 & \textbf{2.7}\tiny$\pm$3.8 & 19.3\tiny$\pm$4.1 & 2.7\tiny$\pm$0.9 & 10.7\tiny$\pm$0.9 & 6.0\tiny$\pm$5.7 & \textbf{22.0}\tiny$\pm$11.8 & \textbf{17.3}\tiny$\pm$5.3 \\
\net{} (WS) & \checkmark & \checkmark & 
    & \textbf{19.3}\tiny$\pm$7.7 & 17.3\tiny$\pm$10.5 & \textbf{39.3}\tiny$\pm$1.9 & 44.7\tiny$\pm$7.4 & 1.3\tiny$\pm$1.9 & 22.0\tiny$\pm$8.6 & 0.7\tiny$\pm$0.9 & 7.3\tiny$\pm$2.5 & 5.3\tiny$\pm$0.9 & 10.7\tiny$\pm$0.9 & 16.8\tiny$\pm$4.3 \\
\net{} (FPS) & \checkmark & \checkmark & 
    & 18.0\tiny$\pm$2.8 & 23.3\tiny$\pm$2.5 & 34.0\tiny$\pm$4.9 & \textbf{51.3}\tiny$\pm$12.7 & 0.7\tiny$\pm$0.9 & 20.0\tiny$\pm$1.6 & 0.0\tiny$\pm$0.0 & 11.3\tiny$\pm$4.1 & 4.7\tiny$\pm$1.9 & 8.0\tiny$\pm$4.3 & 17.1\tiny$\pm$3.6 \\
\net{} (WS) & \checkmark & \checkmark & \checkmark 
    & 15.3\tiny$\pm$7.5 & 24.0\tiny$\pm$7.1 & 30.0\tiny$\pm$7.5 & 39.3\tiny$\pm$5.0 & 0.0\tiny$\pm$0.0 & 18.7\tiny$\pm$8.4 & 0.0\tiny$\pm$0.0 & 11.3\tiny$\pm$6.6 & 8.0\tiny$\pm$3.3 & 9.3\tiny$\pm$2.5 & 15.6\tiny$\pm$4.8 \\
\net{} (FPS) & \checkmark & \checkmark & \checkmark 
    & 15.3\tiny$\pm$7.7 & 21.3\tiny$\pm$6.6 & 36.7\tiny$\pm$5.7 & 34.7\tiny$\pm$10.6 & 0.7\tiny$\pm$0.9 & \textbf{25.3}\tiny$\pm$11.1 & 0.0\tiny$\pm$0.0 & 7.3\tiny$\pm$2.5 & 4.7\tiny$\pm$2.5 & 14.0\tiny$\pm$5.9 & 16.0\tiny$\pm$5.4 \\
\midrule
\multicolumn{15}{c}{\textit{LIBERO-90 Demonstration-level sampling (mix-in)}} \\
All & & & 
    & 24.7\tiny$\pm$9.6 & 8.0\tiny$\pm$3.3 & 14.0\tiny$\pm$4.3 & 51.3\tiny$\pm$9.0 & \textbf{3.3}\tiny$\pm$2.5 & 40.7\tiny$\pm$13.7 & 6.0\tiny$\pm$1.6 & 10.7\tiny$\pm$2.5 & 6.7\tiny$\pm$0.9 & 18.0\tiny$\pm$7.5 & 18.3\tiny$\pm$5.5 \\
DataMIL & \checkmark & \checkmark & \checkmark 
    & 31.3\tiny$\pm$3.4 & 9.3\tiny$\pm$2.5 & \textbf{46.0}\tiny$\pm$9.1 & 59.3\tiny$\pm$12.3 & \textbf{2.7}\tiny$\pm$3.8 & \textbf{56.0}\tiny$\pm$12.8 & 4.0\tiny$\pm$2.8 & 12.7\tiny$\pm$2.5 & \textbf{9.0}\tiny$\pm$1.0 & 22.7\tiny$\pm$4.7 & 25.3\tiny$\pm$5.5 \\
\net{} (WS) & \checkmark & \checkmark & 
    & 27.3\tiny$\pm$8.2 & \textbf{23.5}\tiny$\pm$9.4 & 36.7\tiny$\pm$8.4 & 28.0\tiny$\pm$5.9 & 1.3\tiny$\pm$0.9 & 33.3\tiny$\pm$9.3 & 0.7\tiny$\pm$0.9 & 9.0\tiny$\pm$3.0 & 6.5\tiny$\pm$3.8 & \textbf{24.0}\tiny$\pm$8.5 & 19.0\tiny$\pm$5.8 \\
\net{} (CR) & \checkmark & \checkmark & 
    & \textbf{40.0}\tiny$\pm$9.8 & 18.0\tiny$\pm$13.4 & 40.7\tiny$\pm$15.2 & \textbf{69.3}\tiny$\pm$6.6 & 2.0\tiny$\pm$1.4 & 44.7\tiny$\pm$15.7 & \textbf{12.0}\tiny$\pm$10.7 & \textbf{29.3}\tiny$\pm$13.2 & 6.0\tiny$\pm$2.8 & 10.7\tiny$\pm$1.9 & \textbf{27.3}\tiny$\pm$9.1 \\
\midrule
\multicolumn{15}{c}{\textit{LIBERO-10 Frame-level sampling (50 demos)}} \\
Full (50 demos) & & & 
    & 69.3\tiny$\pm$6.6 & \textbf{88.7}\tiny$\pm$3.4 & 86.0\tiny$\pm$5.9 & 92.7\tiny$\pm$5.0 & 67.3\tiny$\pm$3.4 & \textbf{85.3}\tiny$\pm$7.7 & \textbf{50.0}\tiny$\pm$4.3 & 82.7\tiny$\pm$0.9 & 66.0\tiny$\pm$8.6 & 79.3\tiny$\pm$0.9 & 76.7\tiny$\pm$4.7 \\
\net{} (WS) & \checkmark & \checkmark & 
    & \textbf{80.7}\tiny$\pm$7.5 & 83.3\tiny$\pm$12.7 & \textbf{91.3}\tiny$\pm$0.9 & \textbf{96.0}\tiny$\pm$1.6 & 64.7\tiny$\pm$7.4 & 76.0\tiny$\pm$4.3 & 38.7\tiny$\pm$3.4 & \textbf{88.7}\tiny$\pm$6.8 & \textbf{79.3}\tiny$\pm$5.0 & 79.5\tiny$\pm$1.7 & \textbf{77.8}\tiny$\pm$5.1 \\
\net{} (WS) & \checkmark & \checkmark & \checkmark 
    & 73.3\tiny$\pm$4.1 & 83.3\tiny$\pm$6.2 & 88.7\tiny$\pm$3.8 & 83.3\tiny$\pm$5.0 & \textbf{72.0}\tiny$\pm$9.9 & 81.3\tiny$\pm$6.6 & 34.0\tiny$\pm$8.2 & 87.3\tiny$\pm$4.1 & 70.0\tiny$\pm$7.1 & \textbf{82.0}\tiny$\pm$4.3 & 75.5\tiny$\pm$5.9 \\
\bottomrule
\end{tabular}}
\end{table*}

\subsection{Detailed Performance Correlation Experiments}\label{sec:correlation}
We first define the individual scores used to compare performance in terms of coverage and fidelity, and then discuss dataset-wise results.
Coverage and fidelity scores are computed on the same $D$-dimensional point cloud used for coreset selection ($D=2$ for the main two-axis sampler space, $D=3$ for the (mean, std, third-axis) space): each axis $a = 1, \dots, d$ is independently min--max normalized to $[0,1]$ over the full dataset $P$, and $C \subset P$ denotes the selected coreset. All metrics are oriented so that \emph{higher is better}. For each candidate space, we then compute the Pearson correlation across runs between each score and downstream performance, pooled within coreset ratio, and average it over the scores of each family.
\newpage

\paragraph{Coverage family.}
\begin{itemize}
  \item \textbf{Mean nearest-selected distance:} $\sqrt{d} - \tfrac{1}{N}\sum_{p \in P} \min_{c \in C} \lVert p - c \rVert_2$ (the additive constant $\sqrt{d}$ is the diagonal length of the unit hypercube $[0,1]^d$, so the score stays positive), computed via a $k$-d tree nearest-neighbor query.
  \item \textbf{Coverage@$k$} ($k \in \{3, 10\}$): for every population point, precompute its $k$ nearest neighbors in $P$ (a $k$-d tree query, self included at distance $0$). Coverage@$k$ is the fraction of population points that have \emph{at least one} of their $k$ nearest neighbors in $C$.
  \item \textbf{Recall@$k$} ($k \in \{3, 10\}$): using the same precomputed $k$-NN table, recall@$k$ is the fraction of the full population that appears as a $k$-nearest-neighbor of \emph{at least one} selected point $c \in C$ — how much of $P$ is "reachable" from $C$ within $k$ hops.
  \item \textbf{Tail coverage} (percentiles $\{5, 95\}$): per axis $a$, checks whether $C$ contains at least one point at or below $P$'s 5th percentile and at least one at or above its 95th percentile on that axis; averaged over both indicators and all $D$ axes.
  \item \textbf{Quantile coverage} ($5$ quantile bins per axis, i.e.\ a $5^D$-cell grid): the fraction of population-occupied grid cells that contain at least one selected point.
  \item \textbf{Extreme coverage}: after normalizing $P$ and $C$ to $[0,1]^D$, for each of the $2^d$ corners of the unit hypercube a distance threshold is set to the population's own 10th-percentile distance to that corner; extreme coverage is the fraction of corners for which $C$ has at least one point within that threshold.
\end{itemize}

\paragraph{Fidelity family.}
\begin{itemize}
  \item \textbf{Wasserstein distance} (reported as a similarity): the 1D Wasserstein-1 (earth mover's) distance is computed independently on each axis $a$, averaged over all $D$ axes, and converted to $1 - \overline{W_1}$.
  \item \textbf{Jensen--Shannon similarity}: Gaussian KDEs (SciPy's \texttt{gaussian\_kde}, Scott's-rule bandwidth) are fit to $P$ and to $C$ (subsampled to at most $1500$ points for tractability) and evaluated on a shared grid spanning the population's bounding box ($100$ points/axis for $D=2$, $15$ points/axis for $D=3$); both are normalized to sum to $1$ over the grid, and the Jensen--Shannon divergence $\mathrm{JS}(p \Vert q) = \tfrac12 D_{\mathrm{KL}}(p\Vert m) + \tfrac12 D_{\mathrm{KL}}(q \Vert m)$, $m = \tfrac12(p+q)$, is rescaled to a similarity via $1 - \mathrm{JS}/\ln 2$ (natural-log JS divergence is bounded by $\ln 2$).
  \item \textbf{Density correlation}: the Pearson correlation between the two KDEs' density values on the same grid used for the JS similarity above.
  \item \textbf{Bhattacharyya coefficient}: the discretized Bhattacharyya coefficient $\sum_{\text{grid}} \sqrt{\hat p \cdot \hat q}h^d$ ($h$ = grid spacing) between the same two KDEs, numerically integrated over the shared grid.
\end{itemize}
When an axis is degenerate (e.g.\ a purely 1D selection space) several metrics fall back to a 1D-only comparison of the primary axis, or default to $0$ when no meaningful density estimate can be formed.

We show layerwise results in \tabref{tab:axis-layer}.
Paired with $\delta_{\max}$, the mean of layer $\ell_{12}$ gives the highest fidelity correlation (0.340, against 0.310 for $\ell_{1}$, 0.336 for $\ell_{6}$, and 0.336 for the all-layer mean $\mu_{1:12}$), while $\ell_{6}$ and $\mu_{1:12}$ are slightly ahead on coverage (0.297 and 0.293 against 0.276 for $\ell_{12}$ and 0.267 for $\ell_1$).
We further note that Cityscapes exhibits no correlation to any fidelity metric, highlighting that coverage might be a better metric for this dataset.
We continue our analysis with $\ell_{12}$ as main layer since it outperforms on fidelity and because the difficulty signal in the scores grows with network depth. 

Among the candidates for a second axis next to $\mu_{12}$, $\delta_{\max}$ and $\sigma_{12}$ give the highest fidelity correlation (0.340 and 0.341; \tabref{tab:axis-summary}, per-dataset results in \tabref{tab:axis-second}), followed by $m_{12}$ (0.322) and $\delta_{\mathrm{avg}}$ (0.318). All five are within $0.007$ on coverage. $\delta_{\mathrm{avg}}$ is ahead only on the two depth datasets, so we default to $\delta_{\max}$. 

Inside the sampler space, each of the two axes adds to the other (\tabref{tab:axis-third}): adding $\delta_{\max}$ to $(\mu_{12}, \sigma_{12})$ raises the correlations by 0.028 (coverage) and 0.016 (fidelity), and adding $\sigma_{12}$ to $(\mu_{12}, \delta_{\max})$ raises them by 0.032 and 0.017, on six of seven datasets in both cases.
The three-axis space therefore correlates best with performance on average (0.308 and 0.357, against 0.276--0.280 and 0.340--0.341 for the two-axis subspaces). The differences between options are small, so we treat these axes as a sensible default, not an optimum. 

\begin{table*}[t]
\centering
\small
\setlength{\tabcolsep}{4pt}
\caption{Layer $\ell$ of the mean axis (second axis fixed to $\delta_{\max}$); $\mu_{1:12}$ is the mean over all layers. Pearson correlation between run performance and the metrics of its selected set averaged over the metrics of each family.}
\label{tab:axis-layer}
\begin{tabular}{lcccccccc}
\toprule
& \multicolumn{4}{c}{Coverage $\uparrow$} & \multicolumn{4}{c}{Fidelity $\uparrow$} \\
\cmidrule(lr){2-5} \cmidrule(lr){6-9}
Dataset & $\mu_{1}$ & $\mu_{6}$ & $\mu_{12}$ & $\mu_{1:12}$ & $\mu_{1}$ & $\mu_{6}$ & $\mu_{12}$ & $\mu_{1:12}$ \\
\midrule
CIFAR10 & 0.17 & 0.19 & 0.18 & \textbf{0.20} & 0.26 & 0.27 & \textbf{0.29} & 0.28 \\
PascalVOC & 0.28 & 0.33 & 0.32 & \textbf{0.34} & 0.32 & 0.38 & \textbf{0.40} & \textbf{0.40} \\
Cityscapes & \textbf{0.26} & \textbf{0.26} & \textbf{0.26} & 0.23 & 0.01 & \textbf{0.05} & 0.03 & 0.04 \\
\shortstack[l]{Cityscapes-DVPS (Depth)} & 0.27 & 0.27 & \textbf{0.29} & \textbf{0.29} & 0.48 & \textbf{0.53} & 0.52 & 0.52 \\
\shortstack[l]{Cityscapes-DVPS (Segm.)} & \textbf{0.39} & 0.37 & 0.36 & 0.36 & 0.53 & \textbf{0.54} & \textbf{0.54} & 0.51 \\
\shortstack[l]{SemanticKITTI-DVPS (Depth)} & 0.30 & \textbf{0.45} & 0.33 & 0.43 & 0.34 & \textbf{0.36} & 0.35 & \textbf{0.36} \\
\shortstack[l]{SemanticKITTI-DVPS (Segm.)} & \textbf{0.21} & \textbf{0.21} & 0.20 & 0.20 & 0.23 & 0.23 & \textbf{0.26} & 0.24 \\
\midrule
Mean & 0.267 & \textbf{0.297} & 0.276 & 0.293 & 0.310 & 0.336 & \textbf{0.340} & 0.336 \\
\bottomrule
\end{tabular}
\end{table*}

\begin{table*}[t]
\centering
\small
\setlength{\tabcolsep}{3.5pt}
\caption{Second axis (mean axis fixed to layer $\ell_{12}$; \emph{none} = the mean axis alone). Pearson correlation between run performance and the metrics of its selected set averaged over the metrics of each family.}
\label{tab:axis-second}
\begin{tabular}{lcccccccccc}
\toprule
& \multicolumn{5}{c}{Coverage $\uparrow$} & \multicolumn{5}{c}{Fidelity $\uparrow$} \\
\cmidrule(lr){2-6} \cmidrule(lr){7-11}
Dataset & none & $m_{12}$ & $\sigma_{12}$ & $\delta_{\mathrm{avg}}$ & $\delta_{\max}$ & none & $m_{12}$ & $\sigma_{12}$ & $\delta_{\mathrm{avg}}$ & $\delta_{\max}$ \\
\midrule
CIFAR10 & 0.18 & \textbf{0.19} & \textbf{0.19} & 0.16 & 0.18 & 0.27 & 0.28 & \textbf{0.29} & 0.24 & \textbf{0.29} \\
PascalVOC & \textbf{0.32} & \textbf{0.32} & \textbf{0.32} & 0.28 & \textbf{0.32} & 0.36 & 0.39 & \textbf{0.41} & 0.33 & 0.40 \\
Cityscapes & \textbf{0.38} & 0.29 & 0.27 & 0.19 & 0.26 & 0.02 & \textbf{0.04} & \textbf{0.04} & 0.01 & 0.03 \\
\shortstack[l]{Cityscapes-DVPS (Depth)} & 0.19 & 0.24 & 0.23 & \textbf{0.30} & 0.29 & 0.47 & 0.47 & 0.45 & \textbf{0.53} & 0.52 \\
\shortstack[l]{Cityscapes-DVPS (Segm.)} & \textbf{0.40} & 0.36 & 0.36 & 0.37 & 0.36 & 0.47 & 0.51 & \textbf{0.54} & 0.52 & \textbf{0.54} \\
\shortstack[l]{SemanticKITTI-DVPS (Depth)} & 0.28 & 0.33 & 0.31 & \textbf{0.39} & 0.33 & 0.30 & 0.31 & 0.34 & \textbf{0.35} & \textbf{0.35} \\
\shortstack[l]{SemanticKITTI-DVPS (Segm.)} & 0.16 & 0.23 & \textbf{0.27} & 0.22 & 0.20 & 0.23 & 0.25 & \textbf{0.31} & 0.24 & 0.26 \\
\midrule
Mean & 0.273 & 0.278 & \textbf{0.280} & 0.275 & 0.276 & 0.300 & 0.322 & \textbf{0.341} & 0.318 & 0.340 \\
\bottomrule
\end{tabular}
\end{table*}

\begin{table*}[t]
\centering
\small
\setlength{\tabcolsep}{2pt}
\caption{Correlation between run performance and the metric of its selected set for the three-axis space and for each two-axis subspace. Coverage and fidelity are the same metric families as in the other tables.}
\label{tab:axis-third}
\begin{tabular}{lcccccc}
\toprule
& \multicolumn{3}{c}{Coverage $\uparrow$} & \multicolumn{3}{c}{Fidelity $\uparrow$} \\
\cmidrule(lr){2-4} \cmidrule(lr){5-7}
Dataset & \shortstack{$\mu_{12}$+ $\sigma_{12}$} & \shortstack{$\mu_{12}$+ $\delta_{\max}$} & \shortstack{$\mu_{12}$ + $\sigma_{12}$\\ + $\delta_{\max}$} & \shortstack{$\mu_{12}$+ $\sigma_{12}$} & \shortstack{$\mu_{12}$+ $\delta_{\max}$} & \shortstack{$\mu_{12}$ + $\sigma_{12}$\\ + $\delta_{\max}$} \\
\midrule
CIFAR10 & 0.19 & 0.18 & \textbf{0.20} & 0.29 & 0.29 & \textbf{0.30} \\
PascalVOC & 0.32 & 0.32 & \textbf{0.35} & 0.41 & 0.40 & \textbf{0.42} \\
Cityscapes & 0.27 & 0.26 & \textbf{0.28} & \textbf{0.04} & 0.03 & \textbf{0.04} \\
\shortstack[l]{Cityscapes-DVPS (Depth)} & 0.23 & \textbf{0.29} & \textbf{0.29} & 0.45 & \textbf{0.52} & 0.50 \\
\shortstack[l]{Cityscapes-DVPS (Segm.)} & 0.36 & 0.36 & \textbf{0.42} & 0.54 & 0.54 & \textbf{0.57} \\
\shortstack[l]{SemanticKITTI-DVPS (Depth)} & 0.31 & 0.33 & \textbf{0.37} & 0.34 & 0.35 & \textbf{0.37} \\
\shortstack[l]{SemanticKITTI-DVPS (Segm.)} & \textbf{0.27} & 0.20 & 0.26 & \textbf{0.31} & 0.26 & 0.29 \\
\midrule
Mean & 0.280 & 0.276 & \textbf{0.308} & 0.341 & 0.340 & \textbf{0.357} \\
\bottomrule
\end{tabular}
\end{table*}

\begin{figure}[t]
    \centering
    \begin{subfigure}[b]{\textwidth}
        \centering
        \includegraphics[width=\linewidth]{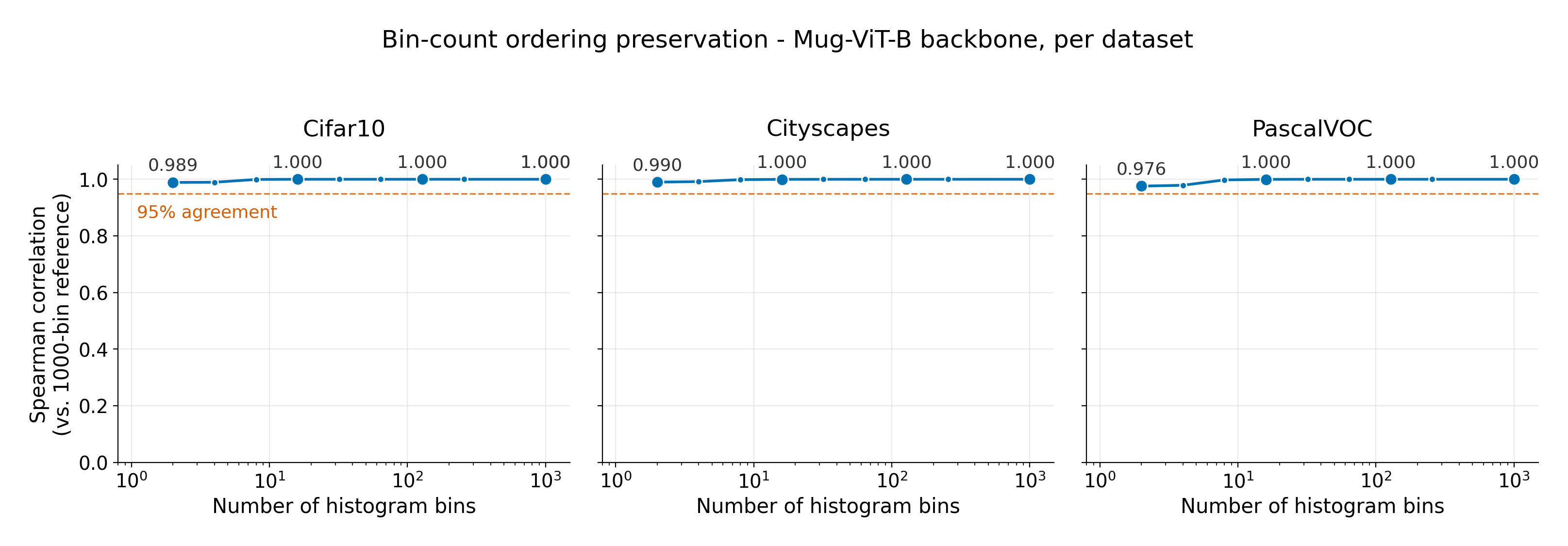}
        \caption{Bin-count ordering preservation, per dataset}
        \label{fig:vdna-bin-comparison}
    \end{subfigure}
        \\[1em]
    \begin{subfigure}[b]{0.48\textwidth}
        \centering
        \includegraphics[width=\linewidth]{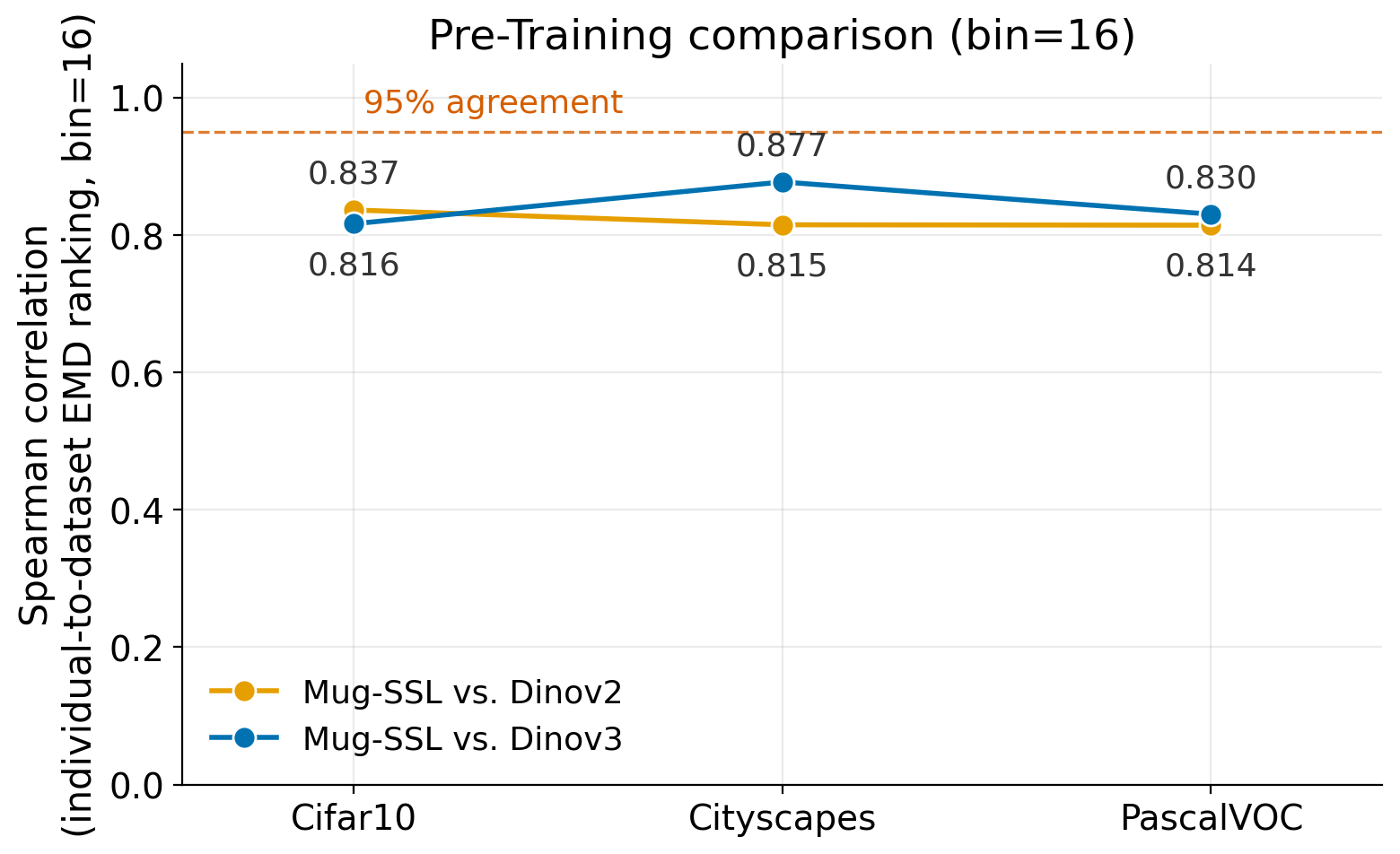}
        \caption{Pre-training comparison (Spearman)}
        \label{fig:vdna-backbone-spearman}
    \end{subfigure}
    \hfill
    \begin{subfigure}[b]{0.48\textwidth}
        \centering
        \includegraphics[width=\linewidth]{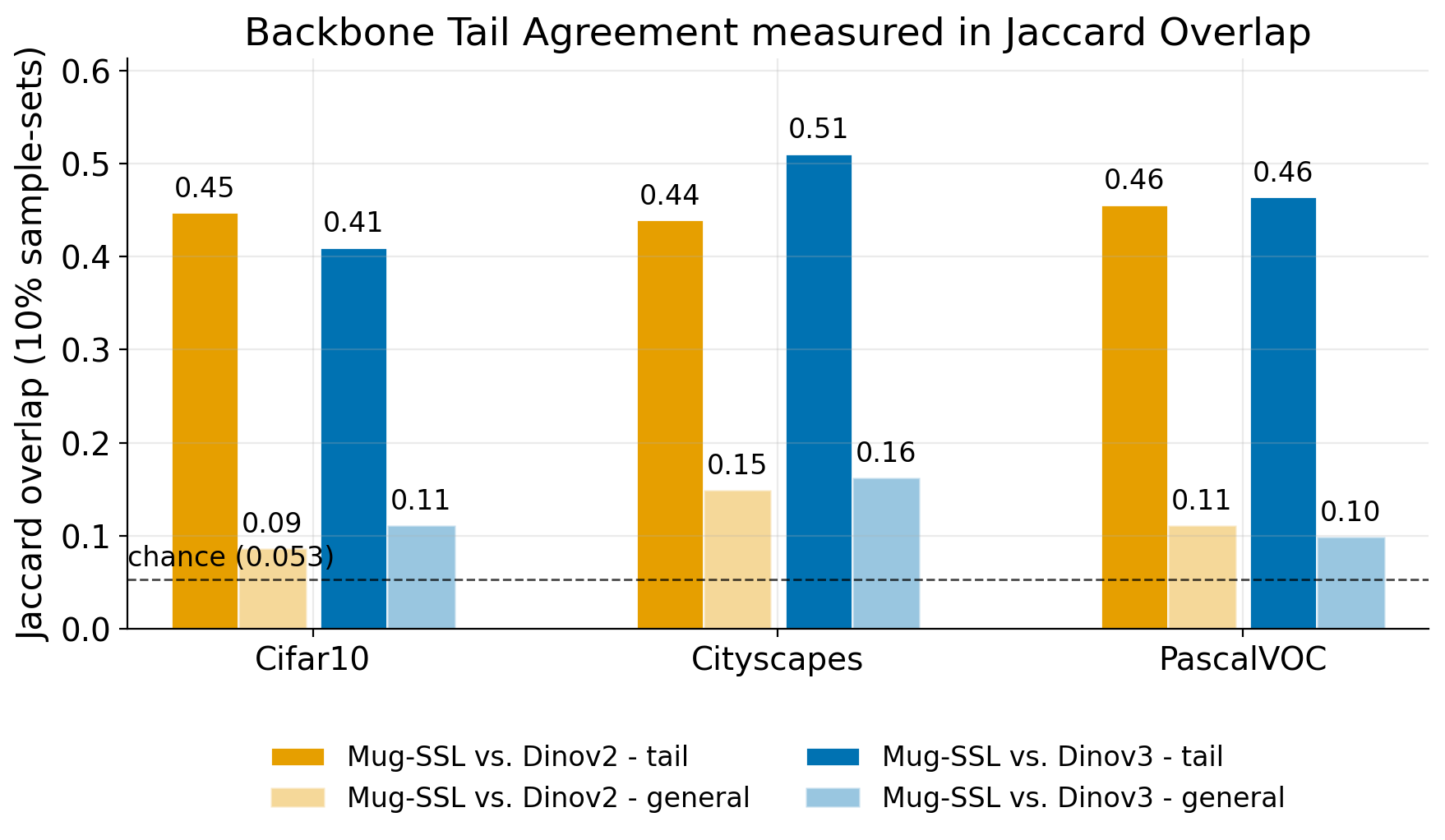}
        \caption{Backbone tail agreement (Jaccard overlap)}
        \label{fig:vdna-backbone-tail}
    \end{subfigure}
    \caption{We compare different histogram bin sizes $B$ in (a) and different backbones for neuron-pooling feature extraction in (b, c).}
    \label{fig:vdna-ablations}
\end{figure}

\subsection{VDNA bins and backbone}\label{sec:vdna_bb}
To evaluate how many histogram bins are needed for the VDNA density score to reliably order samples, we compute the score at bin counts of 2, 16, 128, and 1000 for the Mug-SSL backbone on CIFAR10, Cityscapes, and PascalVOC, and measure the Spearman rank correlation of the resulting per-sample ordering against the 1000-bin reference at each smaller bin count (\figref{fig:vdna-bin-comparison}).
The ordering of samples is confirmed to be identical (Spearman $\rho = 1.000$) from 16 bins up to the 1000-bin reference, across all three datasets. Even 2 bins already preserves the ordering closely ($\rho = 0.976$ to $0.990$). We argue that with the original 1000 bins, a large fraction are empty and do not affect the result, so a substantially smaller bin count can be used with no measurable loss in ordering fidelity. Concretely, on CIFAR10's 50{,}000 training images the materialized representation occupies 1.4GB at 2 bins, 3.1GB at 16 bins, and 11GB at 128 bins. Extrapolating this trend to 1000 bins gives an estimated 28GB for no additional gain in ordering agreement over the 16-bin representation.

We further compare three backbones, Mug-SSL (our default), Dinov2, and Dinov3, normalized in the same
way at 16 bins. We show the Spearman correlation of their full orderings (\figref{fig:vdna-backbone-spearman}) and the Jaccard overlap of the samples
each backbone places in the top and bottom 10\% of the distribution, i.e.\ the dataset's outliers and inliers (\figref{fig:vdna-backbone-tail}).
The backbones' orderings only moderately agree, with Spearman $\rho$ ranging from $0.814$ to $0.877$ across datasets. However, the tails of the distribution overlap much more strongly between backbones: a Jaccard overlap of $0.410$ to $0.511$ in the tails, versus only $0.087$ to $0.163$ for a same-sized band around the median, against a chance level of $0.053$ for two independent random subsets of that size.

\subsection{Ablation over axis choices and Selection Mechanism}\label{sec:axis-choice}
We analyze the effect of our axis space in \tabref{tab:axis_ablation}.
We note that for CIFAR10, choosing $\delta_{\mathrm{avg}}$ instead of $\delta_{\max}$ yields the best performance at $\alpha=0.1$. However, since we seek a universal solution, we stick with $\delta_{\max}$, which we found empirically to correlate best across datasets. For this dataset, using only $\mu_{12}$ already achieves a comparable score, showing that even a single axis captures image-dataset similarity well, whereas using only $\sigma_{12}$ or $\delta_{\max}$ decreases performance by at least $1.23$\,pp. We also test whether using all per-layer means and standard deviations, $\mu_{1}, \dots, \mu_{12}$ and $\sigma_{1}, \dots, \sigma_{12}$, provides a good solution. We observe that this does not achieve the best performance at $\alpha=0.1$. We reason that with small retention rates, the influence of important axes is diluted by redundant axes that do not help. For larger coresets, however, additional axes may be worth exploring, as they enable finer-grained differentiation between near-duplicate samples, consistent with this configuration being only on par with our chosen space at $\alpha=0.5$.

Next, we ablate the sensitivity to the sliced Wasserstein hyperparameters in \tabref{tab:sliced_wasserstein_hparams}. We show that a smaller $g$ only drops performance by $0.34$\,pp at $\alpha=0.1$, whereas a larger $g$ hurts performance more substantially, as it disrupts quantile sampling. Further, $E=10$ is optimal in both settings, with smaller or larger $E$ resulting in drops of at least $0.43$\,pp.


\begin{table}[t]
\centering
\small
\caption{Ablation over axis choices for coreset selection (CIFAR10, sliced Wasserstein selection). Performance is downstream accuracy (\%).}
\label{tab:axis_ablation}
\begin{tabular}{lll cc}
\toprule
& & & \multicolumn{2}{c}{Accuracy $\uparrow$} \\
\cmidrule(lr){4-5}
Mean & Std & Third axis & 0.1 & 0.5 \\
\midrule
$\mu_{1:12}$ & $\sigma_{12}$           & $\delta_{\max}$      & 82.06 & 92.88 \\
$\mu_{1}$              & $\sigma_{12}$           & $\delta_{\max}$         & 80.67 & 93.05 \\
$\mu_{1:12}$   & $\sigma_{1:12}$ & $\delta_{\max}$   & 79.21 & 93.22 \\
$\mu_{1}$              & $\sigma_{1}$            & $\delta_{\max}$         & 79.39 & 93.05 \\
$\mu_{12}$             & $\sigma_{12}$           & $\delta_{\mathrm{avg}}$ & \textbf{82.63} & 93.14 \\
$\mu_{12}$             & -                       & $\delta_{\max}$         & 80.14 & 93.21 \\
$\mu_{12}$             & $\sigma_{12}$           & -                       & 80.95 & 93.30 \\
-                      & $\sigma_{12}$           & -                       & 81.17 & 93.24 \\
-                      & -                       & $\delta_{\max}$         & 80.39 & 93.52 \\
$\mu_{12}$             & -                       & -                       & 82.38 & 93.46 \\
$\mu_{1}, \dots, \mu_{12}$ & $\sigma_{1}, \dots, \sigma_{12}$  & -                   & 81.67 & \textbf{93.58} \\
\midrule
$\mu_{12}$             & $\sigma_{12}$           & $\delta_{\max}$ (main)  & 82.40 & 93.49 \\
\bottomrule
\end{tabular}
\end{table}

\begin{table}[t]
\centering
\small
\caption{Sensitivity to the sliced Wasserstein hyperparameters $E$ (number of projections) and $g$ (near-duplicate gap) on CIFAR10. Performance is downstream accuracy (\%).}
\label{tab:sliced_wasserstein_hparams}
\begin{tabular}{cc cc}
\toprule
& & \multicolumn{2}{c}{Accuracy $\uparrow$} \\
\cmidrule(lr){3-4}
$E$ & $g$ & 0.1 & 0.5 \\
\midrule
3  & 0.01  & 81.35 & 93.34 \\
5  & 0.01  & 80.16 & 93.40 \\
10 & 0.001 & 82.06 & 93.56 \\
10 & 0.01  & 82.40 & 93.49 \\
10 & 0.1   & 80.88 & 93.51 \\
20 & 0.01  & 81.57 & 93.25 \\
\bottomrule
\end{tabular}
\end{table}

\subsection{CIFAR10 Duplication Experiment}\label{sec:cifardub}
To test robustness to redundant data, we construct CIFAR10-Dupl by duplicating all training images ten times and re-evaluate all methods on this dataset under a constant selection amount ($C=5000$ and $C=25000$) in \tabref{tab:cifar10_dupl}. Training-dynamics-based methods collapse dramatically at $C=25000$: AUM, Forgetting, and EL2N drop by $63.36$, $32.50$, and $63.65$pp, respectively, since duplicated samples distort their difficulty estimates and cause them to concentrate selection on a small, redundant subset of hard examples. In contrast, CCS and BOSS, which augment a training-based difficulty signal with a diversity or stratification component, remain stable, as do the fully unsupervised methods (\net{}, ZCore, FreeSel, ELFS), all staying within a few points of their performance on the original dataset. 
This demonstrates that pure training-dynamics-based selection is far more fragile to duplicated data than approaches that account for diversity, whether unsupervised or training-based. This supports our claim that diversity is a more robust criterion for coreset selection than hardness alone.
This robustness is particularly relevant for real-world data collection. Unlike curated benchmarks, video recordings from robots or vehicles naturally contain many near-duplicate frames, making resilience to redundancy a practical necessity rather than a synthetic edge case.

\begin{table}
\centering
\small
\setlength{\tabcolsep}{2pt}
\caption{Downstream accuracy (\%) on CIFAR10, comparing the normal dataset to a version with duplicated samples CIFAR10-Dupl. Values in parentheses show the difference (duplicated $-$ normal).}
\label{tab:cifar10_dupl}
\begin{tabular}{l cc cc}
\toprule
& \multicolumn{2}{c}{Normal $\uparrow$} & \multicolumn{2}{c}{With Duplicates $\uparrow$} \\
\cmidrule(lr){2-3} \cmidrule(lr){4-5}
Method & 50000  ($\alpha=0.1$) & 250000 ($\alpha=0.5$) & 50000 ($\alpha=0.01$) & 250000 ($\alpha=0.05$) \\
\midrule
Full        & 95.05 & 95.05 & 95.33 (+0.28)   & 95.33 (+0.28) \\
\midrule
Random      & 80.90 & 93.11 & 81.46 (+0.56)   & 92.57 ($-$0.54) \\
AUM         & 31.93 & 95.24 & 18.87 ($-$13.06) & 31.88 ($-$63.36) \\
Forgetting  & 39.50 & 95.08 & 36.42 ($-$3.08)  & 62.58 ($-$32.50) \\
EL2N        & 26.31 & 94.90 & 17.64 ($-$8.67)  & 31.25 ($-$63.65) \\
CCS         & 86.49 & 91.16 & 85.90 ($-$0.59)  & 91.28 (+0.12) \\
BOSS        & 84.28 & 94.32 & 84.54 (+0.26)    & 92.03 ($-$2.29) \\
ZCore       & 83.56 & 93.39 & 79.63 ($-$3.93)  & 93.27 ($-$0.12) \\
ELFS AUM    & 86.66 & 89.25 & 86.97 (+0.31)    & 90.98 (+1.73) \\
ELFS Forget & 86.60 & 89.01 & 86.83 (+0.23)    & 91.25 (+2.24) \\
FreeSel     & 82.66 & 93.38 & 81.45 ($-$1.21)  & 92.58 ($-$0.80) \\
\net{} (WS)      & 82.40      &  93.36     & 81.73  ($-$0.67) & 93.07 ($-$0.29)\\
\bottomrule
\end{tabular}
\end{table}

\end{document}